%% file: orvit.tex
\documentclass[10pt,twocolumn,letterpaper]{article}

\usepackage{cvpr}              
\usepackage[accsupp]{axessibility} 

\usepackage{graphicx}
\usepackage{amsmath}
\usepackage{amssymb}
\usepackage{booktabs}

\usepackage[pagebackref,breaklinks,colorlinks]{hyperref}

\usepackage[capitalize]{cleveref}
\crefname{section}{Sec.}{Secs.}
\Crefname{section}{Section}{Sections}
\Crefname{table}{Table}{Tables}
\crefname{table}{Tab.}{Tabs.}

\input{math_commands.tex}

\usepackage{color}
\usepackage{cuted}
\usepackage{capt-of}
\usepackage{tablefootnote}
\usepackage[font=small]{caption} 
\usepackage{multirow} 
\usepackage{footnote}
\usepackage{enumitem}
\usepackage{adjustbox}
\usepackage{subcaption}
\usepackage{caption}
\usepackage{pifont}
\usepackage{algorithm}
\usepackage{algorithmic}
\usepackage{xcolor}
\usepackage{wrapfig}
\usepackage{sidecap}
\definecolor{ForestGreen}{RGB}{34,139,34}
\definecolor{Cerulean}{RGB}{42,82,190}
\definecolor{CornflowerBlue}{RGB}{100,149,237}
\definecolor{Turquoise}{RGB}{48,213,200}
\definecolor{ProcessBlue}{RGB}{0,136,208}
\usepackage{tabularx}

\definecolor{lightgray}{gray}{0.9}
\definecolor{lightblue}{rgb}{0.93,0.95,1.0}
\definecolor{darkgreen}{rgb}{0.0,0.6,0.0}
\definecolor{darkblue}{rgb}{0.0,0.0,0.5}
\definecolor{pinegreen}{rgb}{0.0, 0.47, 0.44}
\definecolor{deepmagenta}{rgb}{0.8, 0.0, 0.8}
\definecolor{amber}{rgb}{1.0, 0.49, 0.0}

\newcommand{\cmark}{\textcolor{darkgreen}{\ding{51}}}
\newcommand{\xmark}{\textcolor{red}{\ding{55}}}

\newcommand{\reals}{\mathbb{R}}

\newcommand{\w}[0]{\texttt{w}}

\newcommand{\ignorebig}[1]{}

\newcommand{\minisection}[1]{\noindent{\textbf{#1}.}}
\newcommand{\tabref}[1]{Table~\ref{#1}}

\newcommand{\figgref}[1]{Figure~\ref{#1}}

\newcommand{\tablestyle}[2]{\setlength{\tabcolsep}{#1}\renewcommand{\arraystretch}{#2}\centering\footnotesize}
\newlength\savewidth

\newcommand{\methodwithboxes}{}
\newcommand{\methodwithoutboxes}{\footnotemark[2]}
\newcommand{\reldif}[1]{$#1\times$}
\newcommand{\ocmodel}{ORViT}
\newcommand{\Mformer}{MF}
\newcommand{\gcol}[1]{{\bf \fontsize{6.5}{42}\selectfont \color{citecolor!80}~(#1)}}

\newcommand{\bcol}[1]{{\fontsize{6.5}{42}\selectfont~(#1) }} 
\definecolor{citecolor}{RGB}{34,139,34}
\definecolor{lightred}{RGB}{241,140,142}
\definecolor{amber(sae/ece)}{rgb}{1.0, 0.49, 0.0}
\definecolor{battleshipgrey}{rgb}{0.52, 0.52, 0.51}
\definecolor{cadmiumorange}{rgb}{0.93, 0.53, 0.18}
\newcommand\orange[1]{\textcolor{amber}{\textbf{#1}}}

\newcommand\darkblue[1]{\textcolor{darkblue}{\textbf{#1}}}

\def\cvprPaperID{7025} 
\def\confName{CVPR}
\def\confYear{2022}

\begin{document}

\title{Object-Region Video Transformers}

\author{
Roei Herzig$^{1}$ \,\,
Elad Ben-Avraham$^{1}$ \,\,
Karttikeya Mangalam$^{2}$ \,\,
Amir Bar$^{1}$ \,\,
Gal Chechik$^{3}$ \,\,\\
Anna Rohrbach$^{2}$ \,\,
Trevor Darrell$^{2}$ \,\,
Amir Globerson$^{1}$ \\ \\
$^1$Tel Aviv University, $^2$UC Berkeley, $^3$Bar-Ilan University, NVIDIA Research}
\vspace{-13pt}
\maketitle

\begin{strip}
    \vspace{-2.5em}
    \centering
    \includegraphics[width=1.0\linewidth]{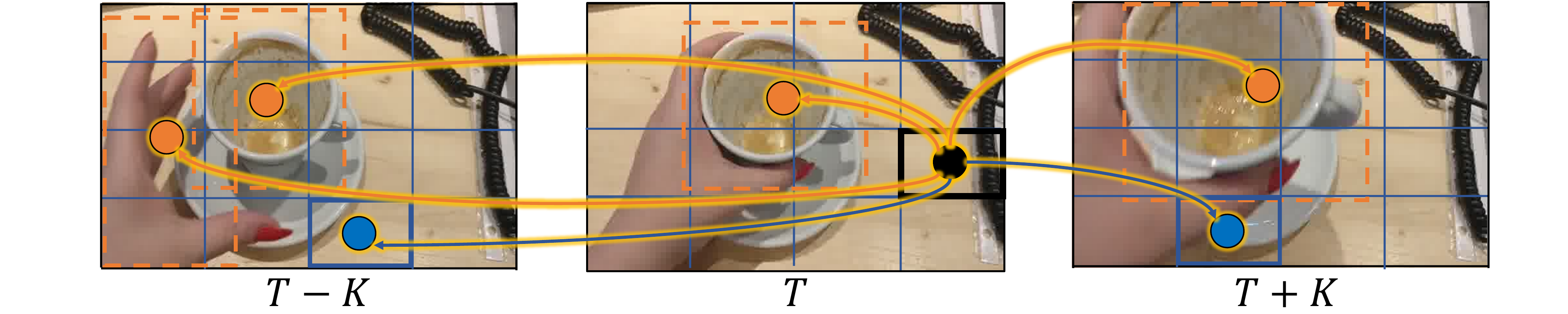}
    \captionof{figure}{Our {\ocmodel} model incorporates object information into video transformer layers. The figure shows the standard (uniformly spaced) transformer patch-tokens in \darkblue{blue}, and object-regions corresponding to detections in \orange{orange}. In {\ocmodel} any temporal patch-token (e.g., the patch in \textbf{black} at time $T$) attends to all patch tokens (\darkblue{blue}) and region tokens (\orange{orange}). This allows the new patch representation to be informed by the objects. Our method shows strong performance improvement on multiple video understanding tasks and datasets, demonstrating the value of a model that incorporates object representations into a transformer architecture.
    }
    \label{fig:teaser}
\end{strip}


\begin{abstract}
\vspace{-1.0em}
%
Recently, video transformers have shown great success in video understanding, exceeding CNN performance; yet existing video transformer models do not explicitly model objects, although objects can be essential for recognizing actions. In this work, we present Object-Region Video Transformers ({\ocmodel}), an \emph{object-centric} approach that extends video transformer layers with a block that directly incorporates object representations. The key idea is to fuse object-centric representations starting from early layers and propagate them into the transformer-layers, thus affecting the spatio-temporal representations throughout the network. Our {\ocmodel} block consists of two object-level streams: appearance and dynamics. In the appearance stream, an ``Object-Region Attention'' module applies self-attention over the patches and \emph{object regions}. In this way, visual object regions interact with uniform patch tokens and enrich them with contextualized object information. We further model object dynamics via a separate ``Object-Dynamics Module'', which captures trajectory interactions, and show how to integrate the two streams. We evaluate our model on four tasks and five datasets: compositional and few-shot action recognition on SomethingElse, spatio-temporal action detection on AVA, and standard action recognition on Something-Something V2, Diving48 and Epic-Kitchen100. We show strong performance improvement across all tasks and datasets considered, demonstrating the value of a model that incorporates object representations into a transformer architecture. For code and pretrained models, visit the project page at \url{https://roeiherz.github.io/ORViT/}
\end{abstract}


\vspace{-1.5em}
\section{Introduction}
\label{sec:intro}

Consider the simple action of ``Picking up a coffee cup'' in~\figgref{fig:teaser}. 
Intuitively, a human recognizing this action would identify the hand, the coffee cup and the coaster, and perceive the upward movement of the cup. This highlights three important cues needed for recognizing actions: what/where are the objects? how do they interact? and how do they move? 
The above perception process allows easy generalization to different compositions of actions. For example, the process of ``picking up a knife" shares some of the components with ``picking up a coffee cup'', namely, the way the object and hand move together. More broadly, representing image semantics using objects facilitates compositional understanding, because many perceptual components remain similar when one object is swapped for another. Thus, a model that captures this compositional aspect potentially requires less examples to train.



It seems intuitively clear that machine vision models should also benefit from such object-focused representations, and indeed this has been explored in the past~\cite{Gupta2007ObjectsIA,Saenko2012MidlevelFI} and more recently by \cite{Wang_videogcnECCV2018,materzynska2019something,arnab2021unified}, who utilize bounding boxes for various video understanding tasks. However, the central question of what is the best way to process objects information still remains. Most object-centric methods to video understanding take a post-processing approach. Namely, they compute object descriptors using a backbone and then re-estimate those based on other objects via message passing or graph networks without propagating the object information into the backbone. Unlike these approaches, we argue that objects should influence the spatio-temporal representations of the scene throughout the network, starting from the early layers (i.e., closer to the input). We claim that self-attention in video transformers is a natural architecture to achieve this result by enabling the attention to incorporate objects {\em as well as} salient image regions.



Video transformers have recently been introduced as powerful video understanding models~\cite{arnab2021vivit,gberta_2021_ICML,mvit2021,patrick2021keeping}, motivated by the success of transformers in language~\cite{devlin-etal-2019-bert} and vision~\cite{detr2020,dosovitskiy2020vit}. In these models, each video frame is divided into patches, and a self-attention architecture obtains a contextualized representation for the patches. However, this approach has no explicit representation of objects. Our key observation is that self-attention can be applied jointly to object representations and spatio-temporal representations, thus offering an elegant and straightforward mechanism to enhance the spatio-temporal representations by the objects.

Motivated by the above, our key goal in this paper is to explicitly fuse object-centric representations into the spatio-temporal representations of video-transformer architectures~\cite{arnab2021vivit}, and do so throughout the model layers, starting from the earlier layers. We propose a general approach for achieving this by adapting the self-attention block~\cite{dosovitskiy2020vit} to incorporate object information. The challenge in building such an architecture is that it should have components for modeling the appearance of objects as they move, the interaction between objects, and the dynamics of the objects (irrespective of their visual appearance). An additional desideratum is that video content outside the objects should not be discarded, as it contains important contextual information. In what follows, we show that a self-attention architecture can be extended to address these aspects. Our key idea is that object regions can be introduced into transformers in a similar way to that of the regular patches, and dynamics can also be integrated into this framework in a natural way. We refer to our model as an ``Object-Region Video Transformer'' (or {\ocmodel}).





We introduce a new {\ocmodel} block, which takes as input bounding boxes and patch tokens (also referred to as spatio-temporal representations) and outputs refined patch tokens based on object information. Within the block, the information is processed by two separate object-level streams: an ``Object-Region Attention'' stream that models appearance and an ``Object-Dynamics Module'' stream that models trajectories.\footnote{Our focus is different from papers on two-stream models in vision, that are not object-centric (see Sec.~\ref{sec:related}).} The appearance stream first extracts descriptors for each object based on the object coordinates and the patch tokens. Next, we append the object descriptors to the patch tokens, and self-attention is applied to all these tokens jointly, thus incorporating object information into the patch tokens (see \figgref{fig:teaser}). The trajectory stream only uses object coordinates to model the geometry of motion and performs self-attention over those. Finally, we re-integrate both streams into a set of refined patch tokens, which have the same dimensionality as the input to our \ocmodel~block. This means that the \ocmodel~block can be called repeatedly. See~\figgref{fig:highlevel} and~\figgref{fig:arch} for visualizations.

We evaluate {\ocmodel} on several challenging video-understanding tasks: compositional and few-shot action recognition on SomethingElse~\cite{materzynska2019something}, where  bounding boxes are given as part of the input; spatio-temporal action detection on AVA~\cite{AVA2018}, where the boxes are obtained via an off-the-shelf detector that was provided by previous methods; and in a standard action recognition task on Something-Something V2~\cite{goyal2017something}, Diving48~\cite{Li_2018_Diving48} and Epic-Kitchen100~\cite{Damen2020RESCALING100EPIC}, where we use class-agnostic boxes from an off-the-shelf detector. Through extensive empirical study, we show that integrating the ORViT block into the video transformer architecture leads to improved results on all tasks. These results confirm our hypothesis that incorporating object representations starting from early layers and propagating them into the spatio-temporal representations throughout the network, leads to better performance.





\section{Related Work}
\label{sec:related}

\minisection{Object-centric models} Recently object-centric models have been successfully applied in many computer vision applications: visual relational reasoning~\cite{battaglia2018relational,zambaldi2018relational,referential_relationships,baradel2018object,herzig2018mapping,Xu2020reason,raboh2020dsg,Jerbi2020LearningOD}, representation learning~\cite{You2020GraphCL}, video relation detection~\cite{Sun2019VideoVR,Liang2019PeekingIT}, vision and language~\cite{Li2019VisualBERTAS,Tan2019LXMERTLC,li2020oscar,Chen2020UNITERUI}, human-object interactions~\cite{Kato2018CompositionalLF,Xu2019LearningTD,Gao2020DRGDR}, and even image generation~\cite{johnson2018image,herzig2019canonical}. The advances and the success of object-centric models in these domains inspired various video-based tasks, such as action localization~\cite{nawhal2021activity}, video synthesis~\cite{2020ActionGraphs}, and action recognition~\cite{zhou2017temporalrelation}. The latter was the focus of varied recent works that designed different object interactions approaches for convolutional models. A line of works~\cite{relnets2017nips, sun2018actor,girdhar2019video} focused on capturing spatial object interactions while ignoring the temporal interactions. STRG~\cite{Wang_videogcnECCV2018} and ORN~\cite{baradel2018object} used spatio-temporal interactions with two consecutive frame interactions, while STAG~\cite{herzig2019stag} considered long-range temporal interaction. Last, Unified~\cite{arnab2021unified} tried to generalize all these models and propose long spatio-temporal object interactions. While all these works focused solely on interactions of visual appearance information, recently STIN~\cite{materzynska2019something} introduced an object-centric model based on object trajectories by modeling bounding box movement. Our {\ocmodel} approach directly combines object appearance, object trajectories, and the overall video, by mapping all computations to spatio-temporal patch tokens. This is particularly natural to do in a transformer framework, as we show here, and results in state-of-the-art performance.


\minisection{Transformers in action recognition} 
Ranging from the early works that employ optical flow based features ~\cite{efros2003recognizing}, to recent transformer based approaches ~\cite{mvit2021}, a wide variety of approaches have been proposed for action recognition. In broad brushstrokes, the proposed approaches have evolved from using temporal pooling for extracting features~\cite{karpathy2014large} to using recurrent networks ~\cite{donahue2015long,yue2015beyond}, through to 3D spatio-temporal kernels~\cite{ji20133d,taylor2010convolutional,tran2015learning,varol2018long, Lin_2019_ICCV, tsn2019wang, carreira2017quo}, and two-stream networks that capture complementary signals (e.g., motion and spatial cues \cite{feichtenhofer2016convolutional,simonyan2014two, slowfast2019}). Unlike these approaches, our work uses two separate object-level streams to leverage object-centric information. In parallel to developments in video understanding, Vision Transformers~\cite{dosovitskiy2020vit, touvron2021training} propose a new approach to image recognition by discarding the convolutional inductive bias entirely and instead employing self-attention operations. Specialized video models such as TimeSformer~\cite{gberta_2021_ICML}, ViViT~\cite{arnab2021vivit}, Mformer (MF)~\cite{patrick2021keeping} and MViT~\cite{mvit2021} form the latest epoch in action recognition models. By generalizing the vision transformers to the temporal domain through the use of spatio-temporal attention, the obtained video transformers are very competitive with their convolutional counterparts both in terms of performance as well scaling behaviour to large data. However, none of the video transformer models leverage object cues, a persistent shortcoming that we aim to address in {\ocmodel}. 

We also note that~\cite{lvu2021} adopts a similar object-centric approach to video understanding. However, our work differs conceptually since \cite{lvu2021} models only the object parts, which is similar to the STRG baselines we consider in the paper. On the other hand, our work introduces objects into transformer layers while keeping the entire spatio-temporal representation. Also,~\cite{lvu2021} pretrains the transformer in a self-supervised fashion on a large dataset (MovieClips), and therefore its empirical results cannot be directly compared to models that are not pretrained in this manner.


\minisection{Spatio-temporal action detection}
The task of action detection requires temporally localizing the action start and end times. A wide variety of methods have been proposed for it, such as actions modeling~\cite{long2019temporalgauss, PGCN2019ICCV, alwassel_2018_actionsearch}, temporal convolutions~\cite{shou2016multistage,lea2017cvpr}, boundaries modeling~\cite{lin2018boundaries,lin2019iccv}, attention~\cite{shi2020weakly,yuan2018marginalized}, structure utilization~\cite{Yuan_2017_CVPR,SSN2017ICCV}, detection based methods~\cite{chao2018cvpr,xu2017rc3d}, end-to-end approaches~\cite{Dai_2017_ICCV, Gao_2017_ICCV,buch2017sst,jain2020actionbytes}, recurrent neural networks~\cite{ma2016lstm,Singh_2016_CVPR,Yeung_2016_CVPR}, and even using language~\cite{richard2016cvpr,Zhukov2019}. Recently, the new MViT~\cite{mvit2021} model showed promising results on action localization in the AVA dataset~\cite{gu2018ava}. However, it does not explicitly model objects, and we show that an {\ocmodel} version of MViT indeed improves performance.

\section{The {\ocmodel} model}
\label{sec:model}
\begin{figure}[t!]
    \centering
    \includegraphics[width=0.95\linewidth]{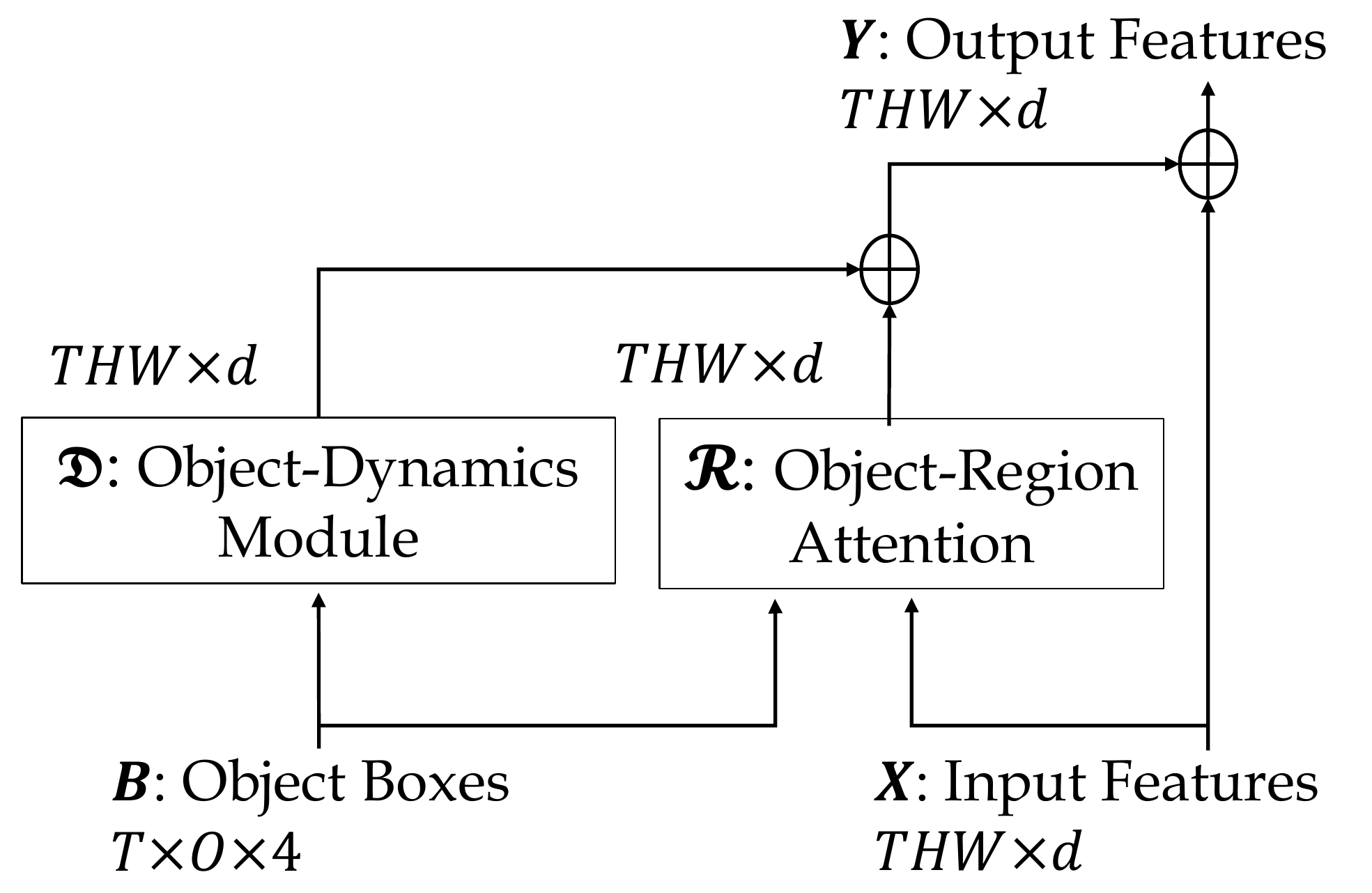}
    \vspace{-.2cm}
    \caption{An \textit{\ocmodel~block}. The input patch-tokens $X$ and boxes $B$ are used as input to the ``Object-Region Attention'' and ``Object-Dynamics Module'' components. Each component outputs a $THW \times d$ tensor and the two tensors are summed to obtain new patch tokens $Y$.}
    \label{fig:highlevel}
    \vspace{-1.0em}
\end{figure}

We next present the Object-Region Video Transformer ({\ocmodel}) model, which explicitly models object appearance and trajectories within the transformer architecture. 
We begin by reviewing the {\vot} architecture, which our model extends upon, in~\Secref{sec:model:vit}, and present {\ocmodel} in \Secref{sec:model:ocmodel}. A high-level overview of {\ocmodel} is shown in~\figgref{fig:highlevel} and detailed in~\figgref{fig:arch}. Briefly, {\ocmodel} repeatedly refines the patch token representations by using information about both the appearance and movement of objects. 


\subsection{The Video Transformer Architecture}
\label{sec:model:vit}

Video transformers~\cite{arnab2021vivit,mvit2021,gberta_2021_ICML} extend the Vision Transformer model to the temporal domain. Similar to vision transformers, the input is first ``patchified'' but with temporally extended 3-D patches instead of 2-D image patches producing a down-sampled tensor X of size $T \times H \times W \times d$. Then, spatio-temporal position embeddings are added  for providing location information. Finally, a classification token (CLS) is appended to $X$, resulting in $THW+1$ tokens in $\reals^d$, to which self-attention are applied repeatedly to produce the final contextualized CLS feature vector.\footnote{In what follows, we omit the count of the CLS feature for brevity.}


\subsection{The \ocmodel~Block}
\label{sec:model:ocmodel}

There are two inputs to the {\ocmodel} block. The first is the output of the preceding transformer block, represented as a set of spatio-temporal tokens  $X\in \reals^{THW \times d}$. The second input is a set of bounding boxes for objects across time\footnote{$O$ represents the maximum number of objects in the training set. If a clip has less than $O$ boxes, we pad the remaining embeddings with zeros.}, denoted by $B\in \reals^{T O\times 4}$. The output of the {\ocmodel} block is a set of refined tokens $Y \in \reals^{THW\times d}$ contextualized with object-centric information. Thus, the {\ocmodel} block can be viewed as a token representation refining mechanism using the object-level information.

As mentioned, we argue that the key cues for recognizing actions in videos are: the objects in the scene, their interactions, and their movement. To capture these cues, we design the \ocmodel~block with the following two object-level streams.
The first stream models the appearance of objects, and their interactions. We refer to it as ``Object-Region Attention'' and denote it by $\mathcal{R}$. The second ``Object-Dynamics Module'' stream (denoted by $\mathcal{D}$) models the interactions between trajectories, independently of their appearance. Importantly, the output of each of the streams is $THW$ token vectors, which can also be interpreted as refined patch representations based on each source of information.

The $\mathcal{D}$ stream only models object dynamics, and thus only uses bounding boxes $B$ as input. We therefore denote its output by $\mathcal{D}(B)$. The stream  $\mathcal{R}$ models appearance and thus depends on both the token representation $X$, and the bounding boxes $B$, and produces $\mathcal{R}(X, B)$.
The final output of the {\ocmodel} block $Y$ is simply formed by the sum of the two streams and an input residual connection:
\begin{equation}
\begin{aligned}
    & Y' := \mathcal{R}(X, B) + \mathcal{D}(B) + X \\
    & Y := Y' + \text{MLP}(\text{LN}(X))
\label{eq:residual1}
\end{aligned}
\end{equation}
%
where $LN$ denotes a LayerNorm operation. 
Next, we elaborate on the two components separately.


\begin{figure}[t!]
    \centering
    \includegraphics[width=.9\linewidth]{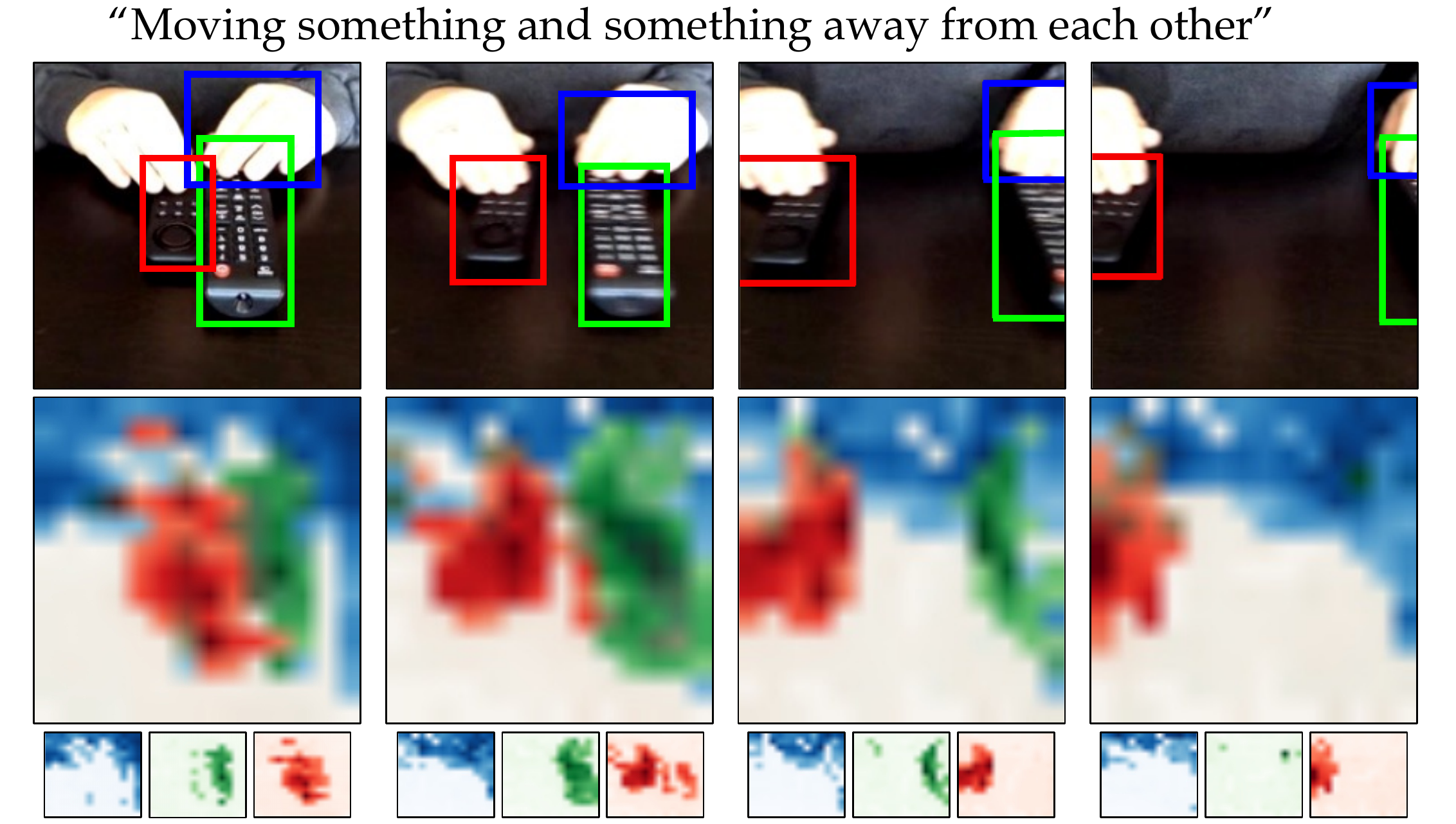}
    \vspace{-.2cm}
    \caption{We visualize the attention allocated to the object tokens in the {\ocmodel} block (red, green, and blue) in each frame of a video describing ``moving two objects away from each other''. It can be seen that each of the two ``remote control'' objects affects the patch-tokens in its region, whereas the hand has a broader map. For more visualizations, please see~\Secref{supp:qual} in supplementary.}
    \label{fig:vis_objs}
    \vspace{-1.0em}
\end{figure}

\begin{figure*}[t!]
    \centering
    \includegraphics[width=.9\linewidth]{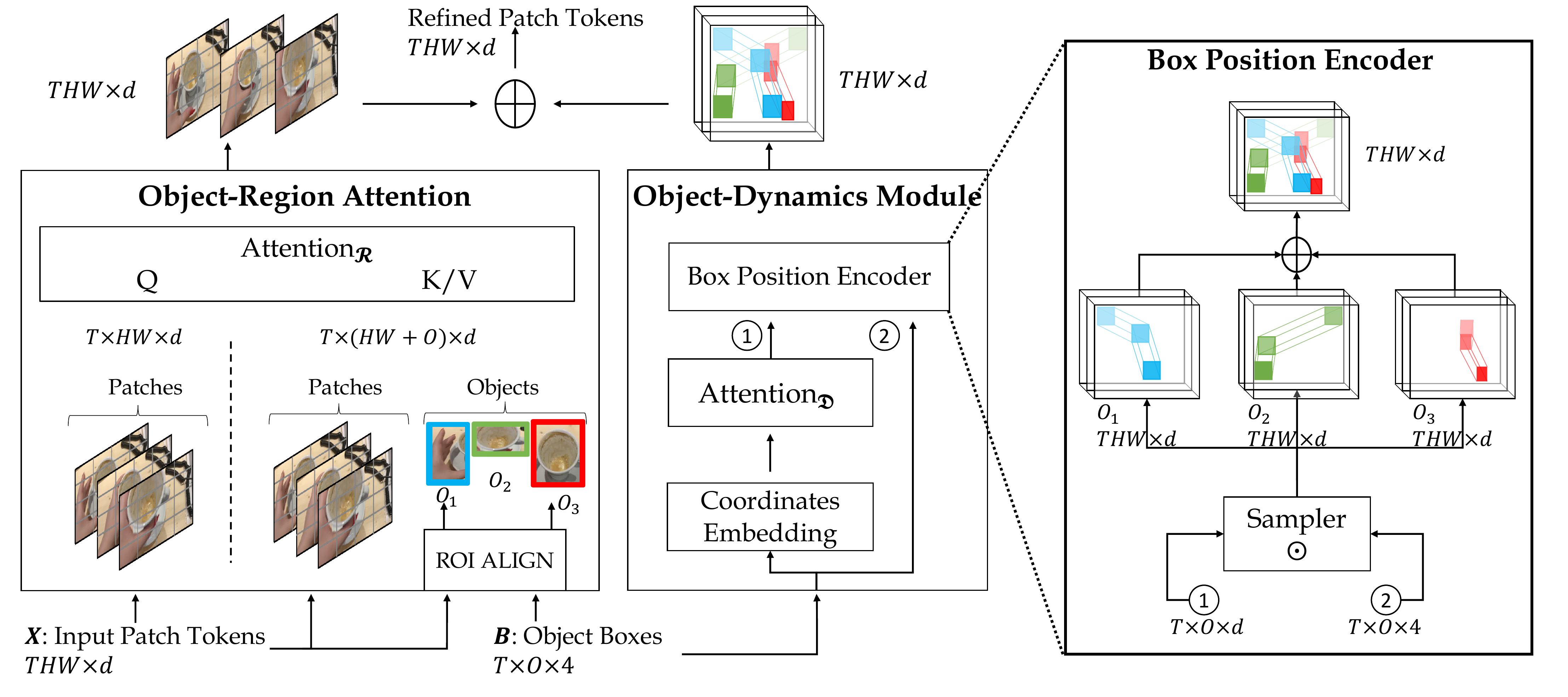}
    \vspace{-0.4cm}
    \caption{\textbf{{\ocmodel} Block architecture}. The block consists of two object-level streams: an ``Object-Region Attention'' that models appearance, and an ``Object-Dynamics Module'' that models trajectories. The two are combined to produce new patch tokens. The ``Box Position Encoder'' maps the output of the trajectory stream to the dimensional of patch tokens. }
    \label{fig:arch}
    \vspace{-1.0em}
\end{figure*}

\minisection{Object-Region Attention} The goal of this module is to extract information about each object and use it to refine the patch tokens. This is done by using the object regions to extract descriptor vectors per region from the input tokens, resulting in $TO$ vectors in $\reals^d$, which we refer to as object tokens. These vectors are then concatenated with the $THW$ patch tokens and serve as the keys and values, while the queries are only the patch tokens. Finally, the output of the block is $THW$ patch tokens. Thus, the key idea is to fuse object-centric information into spatio-temporal representations. Namely, inject the $TO$ object region tokens into $THW$ patch tokens. An overview of our approach is depicted in~\figgref{fig:arch}. We provide further details below.


Given the patch token features $X$ and the boxes $B$, our first goal is to obtain vector descriptors in $\reals^d$ per object and frame. The natural way to do this is via an RoIAlign~\cite{maskrcnn2017iccv} layer, which uses the patch tokens $X$ and box coordinates $B$ to obtain object region crops. This is followed by an MLP and max-pooling to obtain the final object representation in $\reals^d$: 
\begin{equation}
     \mathcal{O} := \text{MaxPool}(\text{MLP}(\text{RoIAlign}(X,B)))
\label{eq:roialign}
\end{equation}

Since this is done per object and per frame, the result is $OT$ vectors in $\reals^d$ (i.e., $\mathcal{O} \in \reals^{TO\times d}$). 
Importantly, this extraction procedure is performed in \emph{each instance of an {\ocmodel} block}, so that it will produce different object tokens at each layer. We also add positional embeddings but leave the details to~\Secref{supp:model:ocmodel} in supplementary.


At this point, we would like to allow the object tokens to refine the patch tokens. We concatenate the object tokens $\mathcal{O}$ with the patch tokens $X$, resulting in $C \in \reals^{T(HW + O) \times d}$. Next $C$ and $X$ are used to obtain queries, keys and values as follows:
\begin{equation}
\begin{aligned}
    & Q := XW_q ~~~~ K := CW_k ~~~~ V := CW_v \\
    & \text{Where~~~~} W_q \text{, } W_k \text{, } W_v \in \reals^{d \times d}
\label{eq:qkv}
\end{aligned}
\end{equation}
Finally, there are several ways to perform spatio-temporal self-attention (e.g., joint and divided attention over space and time, or the recently introduced trajectory attention~\cite{patrick2021keeping}). We use trajectory attention because it performs well empirically. We compare different self-attention versions in~\tabref{supp:tab:abl_appearance} in supplementary. \figgref{fig:vis_objs} also visualizes the ``Object-Region Attention'' learned by our model.

\minisection{Object-Dynamics Module}
To model object dynamics, we introduce a component that only considers the boxes $B$.
We first encode each box via its center coordinate, height and width, and apply an MLP to this vector to obtain a vector in $\reals^d$. Applying this to all boxes results in $\widetilde{L} \in \reals^{TO \times d}$. Next we add a learnable object-time position embedding $\widetilde{P} \in \reals^{TO \times d}$,
 resulting in $\widetilde{B} := \widetilde{L} + \widetilde{P}$. We refer to this as the ``Coordinate Embedding'' step in~\figgref{fig:arch}. Its output can be viewed as $TO$ tokens in $\reals^d$, and we apply self-attention to those as follows: $    \text{Attention}_{\mathcal{D}}(\widetilde{Q}, \widetilde{K}, \widetilde{V}) := \text{Softmax} \left( \frac{\widetilde{Q}\widetilde{K}^T}{\sqrt{d_k}} \right)\widetilde{V}$,
\ignore{

}
where:
    $\widetilde{Q} := \widetilde{B}W_{\widetilde{q}}$,    $\widetilde{K} := \widetilde{B}W_{\widetilde{k}}$, $\widetilde{V} := \widetilde{B}W_{\widetilde{v}}$ and $W_{\widetilde{q}}, W_{\widetilde{k}}, W_{\widetilde{v}} \in \reals^{d \times d}$.
The self-attention output is in $\reals^{TO \times d}$. Next, we would like to transform the objects with a $T \times d$ vector into a spatial volume of $THW \times d$. This is done using the Box Position Encoder described below.

\minisection{Box Position Encoder} The returned features of the {\ocmodel} model should have the same dimensions as the input, namely $THW \times d$. Thus, our main challenge is to project the object embeddings into spatial dimensions, namely $TO \times d$ into $THW \times d$. The naive approach would be to ignore the boxes by expanding every object with vector $T \times d$ into $THW \times d$. However, since the object trajectories contain their space-time location, a potentially better way to do it would consider the object locations. Hence, for each object with corresponding $T \times d$ tokens, we generate a spatial feature $HW \times d$ by placing the object representation vector according to the matching bounding box coordinates using a bilinear interpolation sampler operation~\cite{Jaderberg2015SpatialTN,johnson2018image}.\footnote{Features outside of an object region are set to zeros.} Finally, the output in $HW \times d$ is the sum of all objects from all frames representing the coarse trajectory of the object in spatial dimensions. The process is shown in~\figgref{fig:arch} (right). We show empirically that this approach is better than the naive approach described above.

\subsection{The \ocmodel~model}
\label{sec:model:orvit}

We conclude by explaining how to integrate {\ocmodel} into transformer-based video models. The advantage of {\ocmodel} is that it takes as input the standard spatio-temporal tokens in $\reals^{THW\times d}$ and outputs a refined version of those with the same dimensions. Thus, it acts as a standard transformer layer in terms of input and output, and one can take any transformer and simply add {\ocmodel} layers to it. This is important since it highlights that {\ocmodel} can easily leverage any {\vot} pretrained model, obviating the need for pretraining {\ocmodel}. We experiment with three {\vot} models: TimeSformer~\cite{gberta_2021_ICML}, Mformer {(\Mformer)}~\cite{patrick2021keeping}, and MViT~\cite{mvit2021}. We show that for these, using {\ocmodel} layers improves performance. The only design choice is which layers to apply {\ocmodel} to, while the training methodology remains. We found that it is very important to apply it in early layers while repeated applications keep propagating the information throughout the network. Since the RoIAlign extracts object representations from spatio-temporal representations in each {\ocmodel} layer, multiple {\ocmodel} layers allow the model to consider different object representations throughout the network. In our experiments we apply it in layers $2,7,11$, replacing the original layers without adding depth to the baseline model.



\section{Experiments}
\label{sec:expr}
We evaluate our \textsc{\ocmodel} block on several video understanding benchmarks. Specifically, we consider the following tasks: Compositional Action Recognition (\Secref{expr:comp_fewshot}), Spatio-Temporal Action Detection (\Secref{expr:action_detection}) and Action Recognition (\Secref{expr:act_rec}).

{\minisection{Datasets} We experiment with the following datasets: \textbf{(1) Something-Something v2 (SSv2)}~\cite{goyal2017something} is a dataset containing 174 action categories of common human-object interactions. \textbf{(2) SomethingElse~\cite{materzynska2019something}} exploits the compositional structure of SSv2, where a combination of a verb and a noun defines an action. We follow the official compositional split from~\cite{materzynska2019something}, which assumes the set of noun-verb pairs available for training is disjoint from the set given at test time. \textbf{(3) Atomic Visual Actions (AVA)}~\cite{AVA2018} is a benchmark for human action detection. We report Mean Average Precision (mAP) on AVA-V2.2. \textbf{(4) Epic Kitchens 100 (EK100)}~\cite{Damen2020RESCALING100EPIC} contains 700 egocentric videos of kitchen activities. This dataset includes noun and verb classes, and we report verb, noun, and action accuracy, where the highest-scoring verb and noun pair constitutes an action label. \textbf{(5) Diving48}~\cite{Li_2018_Diving48} contains 48 fine-grained  categories of diving activities.  
}

\minisection{Baselines} In the experiments, we compare {\ocmodel} to several models reported in previous work for the corresponding datasets. These include non-transformer approaches (e.g., \textit{I3D}~\cite{carreira2017quo} and \textit{SlowFast}~\cite{slowfast2019}) as well as state-of-the-art transformers (TimeSformer, Mformer (\Mformer), and MViT). We also cite results for two object-centric models: \textit{STIN}~\cite{materzynska2019something} which uses boxes information, and the Space-Time Region Graph (STRG) model \cite{Wang_videogcnECCV2018} which extracts I3D features for objects and runs a graph neural network on those. Both STIN and STRG use the same input information as {\ocmodel}. Finally, we implement an object-centric transformer baseline combining STRG and STIN: we use the {\Mformer} final patch tokens as input to the STRG model, resulting in STRG feature vector, and concatenate it to the STIN feature vector and the {\Mformer}'s CLS token. We refer to this as \textit{{\Mformer}+STRG+STIN}.

\input{Tables/sota_ssv2_comp_fewshot}


\minisection{Implementation Details}
{\ocmodel} is implemented in PyTorch, and the code will be released in our project page. Our training recipes and code are based on the MViT, {\Mformer}, and TimeSformer code published by the authors. For all tasks and datasets, we use SORT~\cite{Klman1960ANA,Bewley2016_sort} for multi-object tracking to find correspondence between the objects in different frames (no training data is required), see~\Secref{supp:impl:det_tracker} in supp.  We set the number of objects to 4 in SSv2 and EK100, 6 in AVA, and 10 in Diving48. These numbers were chosen by taking the max number of objects per video (as induced by the tracker) across all videos in the training set. 

\vspace{-0.1cm}
\subsection{Compositional \& Few-Shot Action Recognition}
\label{expr:comp_fewshot}
\vspace{-0.1cm}

Several video datasets define an action via a combination of a verb (action) and noun (object). In such cases, it is more challenging to recognize combinations that were not seen during training. This ``compositional'' setting was explored in the ``SomethingElse'' dataset \cite{materzynska2019something}, where verb-noun combinations in the test data do not occur in the training data. The split contains 174 classes with 54,919/54,876 videos for training/validation. This setting is of particular relevance for object-centric models like {\ocmodel}, which can potentially better handle compositional actions. 
The dataset contains annotated bounding boxes that are commonly used as additional input to the model; this allows us to perform a fair comparison against previous methods~\cite{materzynska2019something}. We also evaluate on the few-shot compositional action recognition task in~\cite{materzynska2019something} (see~\Secref{supp:impl:fewshot} in supplementary for details). 


\tabref{tab:comp_fewshot} reports the results for these tasks. {\ocmodel} outperforms all models for both the \textit{Compositional} and \textit{Few-shot} task. Interestingly, the {\Mformer} baseline is relatively strong compared to the previous methods (STRG and STIN). {\ocmodel} provides large improvement over both the previous methods and the baseline {\Mformer} model. We also include results for a strong combined version (MF+STRG+STIN) of the previous methods with {\Mformer} as the backbone.

\input{Tables/sota_ava}

\input{Tables/sota_all_main}

\vspace{-0.1cm}
\subsection{Spatio-temporal Action Detection}
\vspace{-0.1cm}
\label{expr:action_detection}

Next, we evaluate {\ocmodel} on the task of spatio-temporal action detection on the AVA dataset. In the literature, the action detection task on AVA is formulated as a two stage prediction procedure. The first step is the detection of bounding boxes, which are obtained through an off-the-shelf pretrained person detector. The second step involves predicting the action being performed at each detected bounding box. The performance is benchmarked on the end result of these steps and is measured with the Mean Average Precision (MAP) metric. Typically, for fair comparison, the detected person boxes are kept identical across approaches and hence, the final performance depends directly on the ability of the approach to utilize the video and box information.

We follow~\cite{slowfast2019,lfb2019}, using their proposed procedure and the bounding boxes they provided for both {\ocmodel} and evaluation. This enables fair comparison with all previous methods following this standard procedure.\footnote{{Boxes are obtained by pre-training FasterRCNN with a ResNeXt101-FPN~\cite{resnet1,Lin2017FeaturePN} on ImageNet and COCO human keypoint images as in~\cite{slowfast2019}.}}

This task presents an ideal benchmark for evaluating the benefit of {\ocmodel} since all baselines, as well as our model, operate on the same boxes. We trained the MViT-B,16×4 and MViT-B,32×3 models on Kinetics-400~\cite{kay2017kinetics} and report these results. \tabref{tab:sota_ava} shows that \ocmodel-MViT achieves +1.1, +0.7 MAP improvements over the MViT-B 16x4 and MViT-B 32x3, thereby showcasing the power of our proposed object-centric representation fusion scheme.

\vspace{-0.1cm}
\subsection{Action Recognition}
\label{expr:act_rec}



\tabref{table:ar} reports results on the standard action recognition task for several datasets. In contrast to the other tasks presented in this paper, using bounding boxes in action recognition is not part of the task definition. Thus, the comparison should be made carefully while differentiating between non-box and box-based methods. For box-based methods, we consider \textit{STIN}, \textit{STRG} and their combination on top of the same backbone as {\ocmodel}. Next, we explain how the boxes are extracted. For more details about datasets and evaluation see~\Secref{supp:impl} of Supplementary.



\minisection{Box Input to {\ocmodel}} For SSv2, we finetune Faster-RCNN~\cite{nips2015fastercnn} using the annotated boxes as in~\cite{materzynska2019something}. For EK100 and Diving48 we use Faster-RCNN~\cite{nips2015fastercnn} pre-trained on MS~COCO~\cite{Lin2014MSCOCO}. We only use the detector bounding boxes ignoring the object classes. There is no weight sharing between the detector and our model.





\minisection{SSv2} ~\tabref{tab:sota_ssv2} shows that {\ocmodel} outperforms recent methods. The improvement is $1.4\%$ for both \textit{{\Mformer}} and \textit{{\Mformer}-L}, while {\ocmodel} also outperforms other box-based methods, such as \textit{MF+STIN}, \textit{MF+STRG} and their combination. We note that these models do not improve over MF, suggesting that using boxes is non-trivial on large datasets. We also experimented with using manually annotated boxes (as opposed to those obtained by a detector) as an ``oracle'' upper bound to see the potential with annotated box inputs. The results for this oracle evaluation (see~\Secref{supp:exper:results} in supp) are improvements of $7.3\%$ and $6.7\%$ over \textit{{\Mformer}} and \textit{{\Mformer}-L} respectively. This indicates that future improvements in object detectors will benefit object-centric approaches.

\ignore{to see the potential of using object-centric models like {\ocmodel} with annotated box inputs.}



\minisection{Diving48} Here we build {\ocmodel} on top of the TimeSformer model, which was previously reported on this dataset (this demonstrates the ease of adding {\ocmodel} to any transformer model).~\tabref{tab:sota_diving48} shows that our \textit{\ocmodel~TimeSformer} model outperforms the state-of-the-art methods, including TQN~\cite{zhang2021tqn} by a significant margin of $6.2\%$. We obtain an improvement of $8.0\%$ over the baseline TimeSformer model to which {\ocmodel} blocks were added. This again indicates the direct improvement due to the {\ocmodel} block. We note that {\ocmodel} achieves these results using only $32$ frames, significantly less than the previous best results, \textit{TimeSformer-L}, which uses $96$ frames. {\ocmodel} outperforms box-based methods, including \textit{TimeSformer+STIN+STRG} ($4.5\%$), \textit{TimeSformer+STIN} ($7.0\%$), and \textit{TimeSformer+STRG} ($9.9\%$).


\minisection{EK100} ~\tabref{tab:sota_ek100} reports results on EK100. Here we add {\ocmodel} blocks to the \textit{{\Mformer}-HR} model (which is the best performing model on EK100 in \cite{patrick2021keeping}).
Results show that our \textit{\ocmodel~{\Mformer}-HR} model improves the accuracy for all three sub-tasks (with a smaller improvement for nouns). We believe the improvements on EK100 are less impressive than on the other datasets for two main reasons: (a) EK100 is an ego-centric dataset, making the camera movement a significant challenge for our method to model meaningful object interactions. (b) EK100 contains short 2-3 seconds videos, thus temporal reasoning is less effective.



\input{Tables/ablations2}

\subsection{Ablations}


{We perform a comprehensive ablation study on the compositional action recognition task~\cite{materzynska2019something} in the ``SomethingElse'' dataset to test the contribution of the different {\ocmodel} components (\tabref{tab:ablations}). We use the {\Mformer} as the baseline architecture for {\ocmodel}.} For more ablations, see~\Secref{supp:ablations} in supplementary.


\minisection{Components of the {\ocmodel} model} We consider the following versions of our model. (i) Single {\ocmodel} block (no ODM stream\footnote{We refer to the ``Object-Dynamics Module'' as ODM stream.}). We first consider a single application of the {\ocmodel} block, but without the ODM stream. We also compare different {\vot} layers at which to apply our {\ocmodel} block (namely, the {\vot} layer from which to extract the RoI descriptors). We refer to models applied at layer $X$ as \textit{{\ocmodel}[L:X]}. (ii) Single {\ocmodel} block (with ODM stream). Here we augment the single {\ocmodel} block, with the ODM stream. We refer to these models as \textit{{\ocmodel}[L:X]+ODM}. (iii) Multiple {\ocmodel} blocks (with ODM stream). This is the version of {\ocmodel} used in all our experiments. It applies the {\ocmodel} block at multiple layers. We chose layers 2,7 and 11 of the {\vot} model to apply {\ocmodel} block at. All the ablations were performed on the compositional split in SomethingElse. In the ablation table, we refer to this as \textit{{\ocmodel}[L:2,7,11]+ODM}. In the rest of the experiments this is simply referred to as {\ocmodel}.


The results are shown in~\tabref{tab:pitch}. It can be seen that a single {\ocmodel} layer already results in considerable improvement ($66.7\%$), and that it is very important to apply it in the earlier layers rather than at the end. This is in contrast to the current practice in object-centric approaches (e.g., STRG and STIN) that extract RoIs from the final layer. It can also be seen that the ODM stream improves performance (by $2.1\%$ from $66.7\%$ to $68.8\%$). Finally, multiple applications of the layer further improve performance to $69.7\%$.

\minisection{Object-Centric Baselines}
{\ocmodel} proposes an elegant way to integrate {\it object region information} into a {\vot}. Here we consider two other candidate models to achieve this goal. (i) \textit{{\Mformer}+RoIAlign} uses RoIAlign over the last {\vot} layer to extract object features. Then, it concatenates the CLS token with max-pooled object features to classify the action using an MLP. (ii) \textit{{\Mformer}+Boxes} uses coordinates and patch tokens. We use the CLS token from the last layer of {\Mformer}, concatenated with trajectory embeddings. To obtain trajectory embeddings, we use a standard self-attention over the coordinates similar to our ODM stream. The first captures the appearance of objects with global context while the latter captures the trajectory information with global context, both without fusing the object information several times back to the backbone as we do. The results are shown in ~\tabref{tab:abl_objsfusion}. \textit{{\Mformer}+RoIAlign} does not improve over the baseline, while \textit{{\Mformer}+Boxes} improves by $3.5\%$, which is still far from {\ocmodel} ($69.7\%$).


\minisection{How important are the object bounding boxes}
Since {\ocmodel} changes the architecture of the base {\vot} model, we want to check whether the bounding boxes are indeed the source of improvement. We consider several variations where the object bounding boxes are replaced with other values. (i) \textit{All boxes}: all boxes are given the coordinates of the entire image $([0,0,1,1])$. (ii) \textit{Null boxes}: boxes are initialized to zeros. (iii) \textit{Grid boxes}: each of the $4$ bounding boxes is one fourth of the image.
(iv) \textit{Random boxes} - each box is chosen uniformly at random.
See~\tabref{tab:abl_boxes} for results. We observe a large drop in performance for all these baselines, which confirms the important role of the object regions in \ocmodel. Finally, we ask whether tracking information is important, as opposed to just detection. \ignore{Namely, what happens if we keep the bounding boxes, but shuffle their allocation to objects.} We find that this results in degradation from $69.7$ to $68.2$, indicating that the model can perform relatively well with only detection information.


\minisection{Decreasing Model Size} Next, we show that model size can be significantly decreased, incurring a small performance loss. Most the parameters added by {\ocmodel} over the baseline {\Mformer} are in the ODM, and thus it is possible to use a smaller embedding dimension in ODM (see $\widetilde{B}$ in Section~\ref{sec:model:ocmodel}). \tabref{tab:abl_lightweight} reports how the dimension affects the performance, demonstrating that most of the performance gains can be achieved with a model that is \ignore{very} close in size to the original {\Mformer}. More in~\ref{supp:exper:lightweight}$\&$\ref{supp:exper:results} in supp. We would like to highlight that ``Object-Region Attention'' alone (set dimension size to 0; thus ODM is not used) is the main reason for the improvement with only $2\%$ additional parameters.

\section{Discussion and Limitations}

Objects are a key element of human visual perception, but their modeling is still a challenge for machine vision. In this work, we demonstrated the value of an object-centric approach that incorporates object representations starting from early layers and propagates them into the transformer-layers. Through extensive empirical study, we show that integrating the \ocmodel~block into video transformer architecture leads to improved results on four video understanding tasks and five datasets. However, we did not put effort into the object detection and used externally provided boxes, which is a limitation of our work. Replacing the externally provided boxes with boxes that the model generates without supervision will be interesting. We leave this challenge to future work. We believe our approach holds a positive social impact, mainly because it can add compositionality, an essential property for reasoning and common-sense, similar to humans. We do not anticipate a specific negative impact, but we recommend to exercise caution.



\vspace{-10pt}
\subsubsection*{Acknowledgements}
\vspace{-5pt}
We would like to thank Tete Xiao, Medhini Narasimhan, Rodolfo (Rudy) Corona, and Colorado Reed for helpful feedback and discussions. This project has received funding from the European Research Council (ERC) under the European Unions Horizon 2020 research and innovation programme (grant ERC HOLI 819080). Prof. Darrell’s group was supported in part by DoD including DARPA's XAI, and LwLL programs, as well as BAIR's industrial alliance programs. This work was completed in partial fulfillment for the Ph.D. degree of the first author.


{\small
\bibliographystyle{ieee_fullname}
\bibliography{egbib}
}

\newpage
\appendix
\input{supp}

\end{document}

%% file: math_commands.tex
\usepackage{amsmath,amsfonts,bm}

\newcommand{\vot}{video transformer}

\newcommand{\ignore}[1]{}

\def\secref#1{section~\ref{#1}}
\def\Secref#1{Section~\ref{#1}}

\def\eqref#1{equation~\ref{#1}}

\def\1{\bm{1}}

\DeclareMathAlphabet{\mathsfit}{\encodingdefault}{\sfdefault}{m}{sl}
\SetMathAlphabet{\mathsfit}{bold}{\encodingdefault}{\sfdefault}{bx}{n}



%% file: Tables/sota_ssv2_comp_fewshot.tex
\begin{table}[t!]
  \scriptsize
  \tablestyle{2.0pt}{1.00}
   \centering
    \begin{tabular}{l c cc cc cc cc}
            \toprule
            
            \multirow{2}{*}{{Model}} & \multirow{2}{*}{{Boxes}} & \multicolumn{2}{c}{Compositional} & \multicolumn{2}{c}{Base} & \multicolumn{2}{c}{Few-Shot} \\ 
            
             & & {Top-1} & {Top-5} & {Top-1} & {Top-5} & {5-Shot} & {10-Shot}  \\
            
            \midrule
            {I3D~\cite{carreira2017quo}} & \xmark & 42.8 & 71.3 & 73.6 & 92.2 & 21.8 & 26.7 \\
            {SlowFast~\cite{slowfast2019}} & \xmark & 45.2 & 73.4 & 76.1 & 93.4 & 22.4 & 29.2 \\
            {TimeSformer~\cite{gberta_2021_ICML}} & \xmark & 44.2 & 76.8 & 79.5 & 95.6 & 24.6 & 33.8 \\
            {\Mformer~\cite{patrick2021keeping}} & \xmark &  60.2 & 85.8 & 82.8 & 96.2 & 28.9 & 33.8 \\

            \midrule
            
            {STRG (\textbackslash{w}~SF)~\cite{Wang_videogcnECCV2018}} &  \cmark &  52.3 & 78.3 & 75.4 & 92.7 & 24.8 & 29.9 \\
            {STIN} (\textbackslash{w}~SF)~\cite{materzynska2019something} & \cmark &  54.6 & 79.4 & 77.4 & 95.0 & 23.0 & 33.4 \\
            {\Mformer+STRG+STIN} & \cmark &  62.3 & 86.0 & 83.7 & 96.8 & 29.8 & 36.5 \\
            \midrule
            
            \bf {\ocmodel~\Mformer (ours)} & \cmark & \bf {{69.7}} & \bf {{91.0}} & \bf 87.1 & \bf 97.6 & \bf 33.3 & \bf 40.2 \\
            \bottomrule
   \end{tabular}%
    \vspace{-.5em}
    \caption{\textbf{Compositional and Few-Shot Action Recognition} on the ``SomethingElse'' dataset. 
  }
    \label{tab:comp_fewshot}
    \vspace{-1.5em}
\end{table}

%% file: Tables/sota_ava.tex
\begin{table}[t!]
	\centering
	\tablestyle{2.0pt}{1.09}
	\begin{tabular}{l c c c c}
	    \toprule
		{Model} & {Boxes} & {Pretrain} & {mAP} \\ 
		\midrule
		{SlowFast~\cite{slowfast2019}}, $4 \times 16$, R50 & \cmark & {K400} & 21.9 \\ 
		{SlowFast~\cite{slowfast2019}}, $8 \times 8$, R50 & \cmark & {K400} & 22.7 \\ 
		{SlowFast~\cite{slowfast2019}}, $8 \times 8$, R101 & \cmark & {K400} & 23.8 \\ 
		
		\hline
		{{MViT}-B}~\cite{mvit2021}, $16 \times 4$ & \cmark  & {K400} & 25.5 \\ 
		{{MViT}-B}~\cite{mvit2021}, $32 \times 3$ & \cmark  & {K400} & 27.3 \\ 
		\hline
		\bf{\ocmodel~MViT-B, $16 \times 4$ (Ours)} & \cmark & {K400} & \bf 26.6\gcol{$+$1.1} \\
		\bf{\ocmodel~MViT-B, $32 \times 3$ (Ours)} & \cmark & {K400} & \bf 28.0\gcol{$+$0.7} \\
		\midrule
	\end{tabular}
	\vspace{-1.0em}
	\caption{\textbf{Spatio-temporal Action Detection} on AVA-V2.2.
	}
	\label{tab:sota_ava}
	\vspace{-1.5em}
\end{table}

%% file: Tables/sota_all_main.tex
\renewcommand{\thefootnote}{\fnsymbol{footnote}}
\begin{table*}[tb!]
    \begin{subtable}[t]{.35\linewidth}
	\tablestyle{3.5pt}{1.05}
	\scriptsize
	    \centering
        \caption{\bf{Something--Something V2}}
	    \label{tab:sota_ssv2}
		\input{Tables/sota_ssv2_smaller}
	\end{subtable}
    \hfill
    \begin{subtable}[t]{.31\linewidth}
    \tablestyle{0.4pt}{1.05}
    \scriptsize
        \centering
		\caption{\bf{Diving48}}
        \label{tab:sota_diving48}
		\input{Tables/sota_diving48}
	\end{subtable}
	\hfill
	\begin{subtable}[t]{.33\linewidth}
	\tablestyle{1.1pt}{1.05}
	\scriptsize
	    \centering
	    \caption{\bf{Epic-Kitchens100}}
	    \label{tab:sota_ek100}
	    \input{Tables/sota_epic_kitchens}
	\end{subtable}
    \vspace{-0.5em}
    \caption{\textbf{Comparison to state-of-the-art on video action recognition.} We report top-1 ($\%$) and top-5 ($\%$) accuracy on SSv2. On Epic-Kitchens100 (EK100), we report top-1 ($\%$) action (A), verb (V), and noun (N) accuracy. On Diving48 we report top-1 ($\%$). Difference between baselines and {\ocmodel} is denoted by \gcol{$+$X}. IN refers to IN-21K. {We denote methods that do not use bounding boxes with \protect\methodwithoutboxes. For additional results and details, including the model size, see~\secref{supp:exper:results} in supplementary.}}
    \label{table:ar}
    \vspace{-1.0em}
\end{table*}
\renewcommand{\thefootnote}{\arabic{footnote}}

%% file: Tables/sota_ssv2_smaller.tex
\begin{tabularx}
{\linewidth}{@{}l c c c}
    \toprule
    {Model}  & Pretrain & {Top-1} & {Top-5}\\ 
    \midrule
    SlowFast, R101\methodwithoutboxes & K400 & 63.1 & 87.6 \\
    ViViT-L\methodwithoutboxes & {\scriptsize IN+K400}  & 65.4 & 89.8 \\
    MViT-B, 64\methodwithoutboxes & K600  &  68.7 &  91.5 \\

    \midrule
    \Mformer\methodwithoutboxes & {\scriptsize IN+K400} & 66.5 & 90.1 \\
    
    \Mformer+STRG\methodwithboxes & {\scriptsize IN+K400}  & 66.1  & 90.0 \\
    \Mformer+STIN\methodwithboxes & {\scriptsize IN+K400}  & 66.5 & 89.8 \\
    \Mformer+STRG+STIN\methodwithboxes & {\scriptsize IN+K400}  & 66.6  & 90.0 \\
    
    \Mformer-L\methodwithoutboxes  & {\scriptsize IN+K400} & 68.1 & 91.2 \\

    \midrule
    
    \bf \ocmodel~\Mformer\methodwithboxes (Ours) & {\scriptsize IN+K400}  & \textbf{67.9}\gcol{$+$1.4} & 90.5\gcol{$+$0.4} \\
    
    
    \bf \ocmodel~\Mformer-L\methodwithboxes (Ours) & {\scriptsize IN+K400}  & \bf 69.5\gcol{$+$1.4} & \bf 91.5\gcol{$+$0.3} \\
    
    \bottomrule
\end{tabularx}

%% file: Tables/sota_diving48.tex
\renewcommand*{\thefootnote}{\fnsymbol{footnote}}
\begin{tabularx}
{\linewidth}{@{}l c c c c}
    \toprule
    {Model} & {Pretrain~} & {Frames} & {Top-1} \\ 
    \midrule

    SlowFast, R101\methodwithoutboxes & K400 & 16 & 77.6 \\
    TimeSformer\methodwithoutboxes & IN & 16 & 74.9 \\
    TimeSformer-L\methodwithoutboxes & IN & 96 & 81.0 \\
    TQN\methodwithoutboxes & K400 & ALL & 81.8  \\
    \midrule
    TimeSformer\methodwithoutboxes & IN & 32 & 80.0 \\
    TimeSformer + STRG\methodwithboxes  & IN & 32 & 78.1 \\
    TimeSformer + STIN\methodwithboxes  & IN & 32 & 81.0 \\
    TimeSformer + STRG + STIN\methodwithboxes  & IN & 32 & 83.5 \\
    \midrule
    \multirow{2}{*}{\bf \ocmodel~TimeSformer\methodwithboxes(Ours)} & \multirow{2}{*}{\scriptsize IN} & \multirow{2}{*}{32} & \bf 88.0 \\
     & & & \bf \gcol{$+$8.0} \\
    \bottomrule
\end{tabularx}

%% file: Tables/sota_epic_kitchens.tex
\renewcommand*{\thefootnote}{\fnsymbol{footnote}}
\begin{tabularx}
    {\linewidth}{@{}l c c c c}
    \toprule
    {Method} & Pretrain & {A} & {V} & {N}  \\ 
    \midrule
    SlowFast, R50\methodwithoutboxes & K400  & 38.5 & 65.6 & 50.0 \\
    
    ViViT-L\methodwithoutboxes & {\scriptsize IN+K400}  & 44.0 & 66.4 & 56.8 \\
    \Mformer\methodwithoutboxes & {\scriptsize IN+K400} & 43.1 & 66.7 & 56.5 \\
    \Mformer-L\methodwithoutboxes & {\scriptsize IN+K400} & 44.1 & 67.1 & 57.6 \\
    
    \midrule
    \Mformer-HR\methodwithoutboxes & {\scriptsize IN+K400} & 44.5 & 67.0 & 58.5 \\
    MF-HR + STRG\methodwithboxes & {\scriptsize IN+K400}  & 42.5 & 65.8 & 55.4 \\
    MF-HR + STIN\methodwithboxes & {\scriptsize IN+K400}  & 44.2 & 67.0 & 57.9 \\
    MF-HR + STRG + STIN\methodwithboxes & {\scriptsize IN+K400}  & 44.1 & 66.9 & 57.8 \\
    \midrule
    \multirow{2}{*}{\bf \ocmodel~\Mformer-HR\methodwithboxes (Ours)}  & \multirow{2}{*}{\scriptsize IN+K400} & \bf \scriptsize 45.7 & \bf \scriptsize 68.4 & \bf \scriptsize 58.7 \\
    & & \bf \scriptsize \gcol{$+$1.2} & \bf \scriptsize \gcol{$+$1.4} & \bf \scriptsize \gcol{$+$.2} \\
    \bottomrule
\end{tabularx}


%% file: Tables/ablations2.tex
\begin{table*}[tb]
    \begin{subtable}[t]{.33\linewidth}
    \tablestyle{1.0pt}{1.00}
    \scriptsize
        \centering
		\caption{\bf{Components}}
		\label{tab:pitch}
		\input{Tables/abl_pitch}
	\end{subtable}
	\begin{subtable}[t]{.19\linewidth}
    \tablestyle{3.5pt}{1.00}
    \scriptsize
        \centering
		\caption{\bf{Object-centric Baselines}}
		\label{tab:abl_objsfusion}
		\input{Tables/abl_objsfusion}
	\end{subtable}
    \begin{subtable}[t]{.22\linewidth}
    \tablestyle{1.5pt}{1.00}
    \scriptsize
        \centering
		\caption{\bf{Boxes}}
		\label{tab:abl_boxes}
		\input{Tables/abl_boxes}
	\end{subtable}
	\begin{subtable}[t]{.24\linewidth}
    \tablestyle{2.0pt}{1.00}
    \scriptsize
        \centering
		\caption{\bf{ODM Dimension}}
		\label{tab:abl_lightweight}
		\input{Tables/abl_lightweight}
	\end{subtable}
	\vspace{-.5em}
    \caption{\textbf{Albations.} We report top-1 and top-5 action accuracy on the SomethingElse split. We show (a) Contribution of {\ocmodel} components (with parameters number in $10^6$ and GFLOPS in $10^9$). (b) Other Object-centric baselines. (c) {\ocmodel} with different boxes input, and (d) The effect of ``Object-Dynamics Module'' (ODM) embedding dimension. More ablations are in~\Secref{supp:ablations} in supplementary.}
    \label{tab:ablations}
    \vspace{-1.0em}
\end{table*}

%% file: Tables/abl_pitch.tex



\begin{tabularx}
{\linewidth}{@{}l c c c c}
    \toprule
    {Layers} & {Top-1} & {Top-5} & GFLOP & Param \\ 
    \midrule
    MF & 60.2 & 85.8 & \reldif~1(370) & \reldif~1(109)  \\ 
	{\ocmodel} [L:12] & 63.9 & 87.6 & \reldif~1.01 & \reldif~1.01 \\ 
	{\ocmodel} [L:2] & 66.7 & 89.2 & \reldif~1.01 & \reldif~1.01 \\ 
	{\ocmodel} [L:2]+ODM & 68.8 & 90.5 & \reldif~1.03 & \reldif~1.12 \\ 
	{\ocmodel} [L:2,7,11]+ODM & \textbf{69.7} & \textbf{91.0} & \reldif~1.09 & \reldif~1.36 \\ 
    \bottomrule
\end{tabularx}


%% file: Tables/abl_objsfusion.tex
\begin{tabularx}
{\linewidth}{@{}l c c r@{} r}
    \toprule
    ~~{Model} & {Top-1} & {Top-5} \\ 
    \midrule
    ~~MF & 60.2 & 85.8 \\
    ~~MF + RoIAlign & 59.6 & 84.5 \\
    ~~MF + Boxes & 63.7 & 86.9 \\
    
    ~~\bf \ocmodel~(Ours) & \bf 69.7 & \bf 91.0 \\
    \bottomrule
\end{tabularx}

%% file: Tables/abl_boxes.tex
\begin{tabularx}
{\linewidth}{@{}l c c r@{} r}
    \toprule
    {Model} & {Top-1} & {Top-5} \\ 
    \midrule
    Full boxes & 60.9 & 84.5 \\
    Null boxes & 60.4 & 84.2 \\
    Grid boxes & 60.9 & 84.8 \\
    Random boxes & 60.7 & 85.0 \\
    \bf Object Regions (Ours) & \bf 69.7 & \bf 91.0 \\
    \bottomrule
\end{tabularx}

%% file: Tables/abl_lightweight.tex
\begin{tabularx}
{\linewidth}{@{}l c c c c}
    \toprule
    {Dim.} & {Top-1} & {Top-5} & GFLOP & Param \\ 
    \midrule
    MF & 60.2 & 85.8 & \reldif~1 (370) & \reldif~1 (109) \\ 
	0 & 67.4 & 89.8 & \reldif~1.03 & \reldif~1.02 \\ 
	128 & 68.7 & 90.3 & \reldif~1.03 & \reldif~1.03 \\ 
	256 & 68.9 & 90.5 & \reldif~1.03 & \reldif~1.05 \\ 
	768 & 69.7 & 91.0 & \reldif~1.1 & \reldif~1.36 \\ 
    \bottomrule
\end{tabularx}


%% file: supp.tex
\renewcommand{\thefootnote}{\fnsymbol{footnote}}
\clearpage
\section*{Supplementary Material}

In this supplementary file, we provide additional information about our model, implementation details, experimental results, and qualitative examples. Specifically,~\Secref{supp:impl} provides additional implementation details,~\Secref{supp:model} provides additional model details, ~\Secref{supp:ablations} provides additional ablations of our approach,~\Secref{supp:exper} provides more experiment results, and we show qualitative visualizations to demonstrate our approach in~\Secref{supp:qual}.


\section{Additional Implementation Details}
\label{supp:impl}
We add {\ocmodel} to multiple existing transformer models Mformer (\Mformer)~\cite{patrick2021keeping}, MViT~\cite{mvit2021}, and TimesFormer~\cite{gberta_2021_ICML}. These are all implemented based on the SlowFast~\cite{slowfast2019} library (available at \url{https://github.com/facebookresearch/SlowFast}), and we implement {\ocmodel} based on this repository. Next, we elaborate on how we extract the object regions, and for each dataset, we add additional implementation details.

\subsection{Detector and Tracker} 
\label{supp:impl:det_tracker}
\minisection{Detector} In all action recognition datasets we used Faster R-CNN detector~\cite{nips2015fastercnn,maskrcnn2017iccv} with ResNet-50 backbone~\cite{resnet1} and Feature Pyramid Network (FPN)~\cite{Lin2017FeaturePN} that is pre-trained on the MS~COCO~\cite{Lin2014MSCOCO} dataset. We used the Detectron2~\cite{wu2019detectron2} implementation. In SSv2, the detector is finetuned using the bounding boxes annotation. During finetuning the class categories are \textit{hand} and \textit{object}. For AVA, we used the provided detection boxes for the spatio-temporal action detection task that were first obtained by Faster-RCNN pre-trained over MS~COCO and then fine-tuned on AVA, as in~\cite{slowfast2019,lfb2019}. We set the number of objects in our model to 4 in SSv2 and EK100, 6 in AVA, and 10 in Diving48. If fewer objects are presented, we set the object coordinates with a zero vector. These numbers were chosen by taking the max number of objects (after removing the outliers) per video (as induced by the tracker output) across all videos in the training set.

\minisection{Tracker} Once we have the detector results, we apply multi-object tracking to find correspondence between the objects in different frames. We use SORT~\cite{Bewley2016_sort}: a simple tracker implemented based on Kalman Filter~\cite{Klman1960ANA} and the Hungarian matching algorithm (KM)~\cite{Kuhn1955TheHM}. At each step, the Kalman Filter predicts plausible instances in the current frame based on previous tracks. Next, the predictions are matched with single-frame detections by the Hungarian matching algorithm. It is important to note that the tracker does not require any training and does not use any additional data. If an object does not appear in one of the frames, we set the coordinates in these frames to zeros. We provide some stats about the accuracy of the tracking step. Based on two consecutive frames of SSv2, SORT yields an 90.1\% exact match between boxes.

\minisection{Choosing the $O$ on each dataset} We mentioned that $O$ is set to the maximum number of objects per video (generated by the tracker) across all videos in the training set. In addition, we also tested an approach that does not require setting $O$ ahead of time: for each batch take the maximum number of objects in any clip in the batch, and pad all clips to this number. This results in a very similar performance (-0.1\%). This padding approach reduces the pre-processing step of finding the numbers for any dataset by choosing a fixed and large enough number. We also measured the inference runtime (milliseconds per clip) with and without SORT, and it increased by x1.18 (from 60.9ms to 71.6ms). Additionally, the object tokens add x1.03 FLOPS and x1.05 run-time.


\subsection{Something-Something v2}
\label{supp:impl:ssv2}
\minisection{Dataset} The SomethingElse~\cite{goyal2017something} contains 174 action categories of common human-object interactions. We follow the official splits from~\cite{goyal2017something}. 

\minisection{Optimization details} For the standard SSv2~\cite{materzynska2019something} dataset, we trained $16$ frames with sample rate $4$ and batch-size $48$ on $8$ RTX 3090 GPUs. We train our network for 35 epochs with Adam optimizer~\cite{kingma2014adam} with a momentum of $0.9$ and Gamma $0.1$. Following~\cite{patrick2021keeping}, we use $lr = 5e-5$ with $\times10$ decay steps at epochs $0, 20, 30$. Additionally, we used Automatic Mixed Precision, which is implemented by PyTorch. We initialize from a Kinetics-400 pre-trained model~\cite{kay2017kinetics}. For the \textit{\ocmodel~\Mformer-L} model, we fine-tuned from the SSv2 pre-trained model provided by~\cite{patrick2021keeping} and train with $32$ frames. The optimization policy is similar to the above, except we used a different learning rate: $1e-5$ for the pre-trained parameters, and $1e-4$ for the {\ocmodel} parameters. 

For the compositional action recognition task, we trained on the SomethingElse splits~\cite{materzynska2019something}. We train with a batch size of $32$ and a learning rate of $3e-5$.

\minisection{Regularization details} We use weight decay of $0.05$, a dropout~\cite{Hinton2012dropout} of $0.5$ before the final classification, dropout of $0.3$ after the \ocmodel~block, and DropConnect~\cite{Huang2016deepstochastic} with rate $0.2$.

\minisection{Training details} We use a standard crop size of $224$, and we jitter the scales from $256$ to $320$. Additionally, we use RandAugment~\cite{NEURIPS2020Randaugment} with maximum magnitude $20$ for each frame separately.

\minisection{Inference details} We take 3 spatial crops per single clip to form predictions over a single video in testing as done in ~\cite{patrick2021keeping}.




\subsection{EpicKitchens100}
\label{supp:impl:ek100}

\minisection{Dataset} EK100~\cite{Damen2020RESCALING100EPIC} contains 700 egocentric videos of daily kitchen activities. This dataset includes 300 noun and 97 verb classes, and we report verb, noun, and action top-1 accuracy, while the highest-scoring of the verb and noun pairs constitutes the action label.

\minisection{Optimization details} We trained over videos of $16$ frames with sample rate $4$ and batch-size $16$ on $8$ Quadro RTX 8000 GPUs. We train our network for 35 epochs with Adam optimizer~\cite{kingma2014adam} with a momentum of $0.9$ and Gamma $0.1$. Following~\cite{patrick2021keeping}, we use $lr = 5e-5$ with $\times10$ decay steps at epochs $0, 30, 40$. Additionally, we used Automatic Mixed Precision, which is implemented by PyTorch. We initialize from a Kinetics-400 pre-trained model~\cite{kay2017kinetics}.

\minisection{Training details} We use crop size of $336$ for the \textit{\ocmodel~\Mformer-HR}. We jitter the scales from $384$ to $480$. Additionally, we use RandAugment~\cite{NEURIPS2020Randaugment} with maximum magnitude $20$.

\minisection{Inference details} We take 3 spatial crops with 10 different clips sampled randomly to aggregate predictions over a single video in testing.

\subsection{Diving48}
\label{supp:impl:diving48}

\minisection{Dataset} Diving48~\cite{Li_2018_Diving48} contains 16K training and 3K testing videos spanning 48 fine-grained diving categories of diving activities. For all of these datasets, we use standard classification accuracy as our main performance metric.

\minisection{Optimization details} We trained over videos of $32$ frames with sample rate $8$ and batch-size $8$ on $8$ Quadro RTX 8000 GPUs. We train our network for 35 epochs with Adam optimizer~\cite{kingma2014adam} with a momentum of $0.9$ and Gamma $0.1$. We use $lr = 3.75e-5$ with $\times10$ decay steps at epochs $0, 20, 30$. Additionally, we used Automatic Mixed Precision, which is implemented by PyTorch. We initialize from a Kinetics-400 pre-trained model~\cite{kay2017kinetics}.

\minisection{Training details} We use a standard crop size of $224$ for the standard model and jitter the scales from $256$ to $320$. Additionally, we use RandomFlip augmentation. Finally, we sampled the $T$ frames from the start and end diving annotation time, followed by~\cite{zhang2021tqn}. 

\minisection{Inference details} We take 3 spatial crops per single clip to form predictions over a single video in testing same as in~\cite{gberta_2021_ICML}.

\minisection{Object-Dynamics Module} As we show in~\tabref{supp:tab:abl_motion}, we compared different self-attention mechanisms, and the standard self-attention usually performed better. However, we observe a slight improvement when we perform a trajectory self-attention~\cite{patrick2021keeping} instead of the standard self-attention.

\subsection{AVA}
\label{supp:impl:ava}

\minisection{Architecture} SlowFast~\cite{slowfast2019} and MViT~\cite{mvit2021} are using a detection architecture with a RoI Align head on top of the spatio-temporal features. We followed their implementation to allow a direct comparison. Next we elaborate on the RoI Align head proposed in SlowFast~\cite{slowfast2019}. First, we extract the feature maps from our \ocmodel~MViT model by using the RoIAlign layer. Next, we take the 2D proposal at a frame into a 3D RoI by replicating it along the temporal axis, followed by a temporal global average pooling. Then, we max-pooled the RoI features and fed them to an MLP classifier for prediction. 

\minisection{Optimization details} To allow a direct comparison, we used the same configuration as in MViT~\cite{mvit2021}. We trained $16$ frames with sample rate $4$, depth of $16$ layers and batch-size $32$ on $8$ RTX 3090 GPUs. We train our network for 30 epochs with an SGD optimizer. We use $lr = 0.03$ with a weight decay of $1e-8$ and a half-period cosine schedule of learning rate decaying. We use mixed precision and fine-tune from an MViT-B, $16\times4$ pre-trained model.



\minisection{Training details} We use a standard crop size of $224$ and we jitter the scales from $256$ to $320$. We use the same ground-truth boxes and proposals that overlap with ground-truth boxes by $IoU > 0.9$ as in~\cite{slowfast2019}.

\minisection{Inference details} We perform inference on a single clip with $16$ frames. For each sample, the evaluation frame is centered in frame $8$. We use a crop size of $224$ in test time. We take 1 spatial crop with 10 different clips sampled randomly to aggregate predictions over a single video in testing.

\subsection{Few Shot Compositional Action Recognition}
\label{supp:impl:fewshot}

We also evaluate on the few-shot compositional action recognition task in~\cite{materzynska2019something}. For this setting, we have 88 \textit{base} action categories and 86 \textit{novel} action categories. We train on the base categories (113K/12K for training/validation) and finetune on few-shot samples from the novel categories (for 5-shot, 430/50K for training/validation; for 10-shot, 860/44K for training/validation).

\subsection{SomethingElse}
\label{supp:impl:smthelse}
\minisection{Dataset} The SomethingElse~\cite{materzynska2019something} contains 174 action categories with 54,919 training and 57,876 validation samples. The proposed compositional~\cite{materzynska2019something} split in this dataset provides disjoint combinations of a verb (action) and noun (object) in the training and testing sets. This split defines two disjoint groups of nouns $\{\mathcal{A}, \mathcal{B}\}$ and verbs $\{1, 2\}$. Given the splits of groups, they combine the training set as $1\mathcal{A} + 2\mathcal{B}$, while the validation set is constructed by flipping the combination into $1\mathcal{B} + 2\mathcal{A}$. In this way, different combinations of verbs and nouns are divided into training or testing splits.


\input{Tables/ablations_supp}

\minisection{Optimization details} We trained $16$ frames with sample rate $4$ and batch-size $32$ on $8$ RTX 3090 GPUs. We train our network for 35 epochs with Adam optimizer~\cite{kingma2014adam} with a momentum of $0.9$ and Gamma $0.1$. We use $lr = 3e-5$ with $\times10$ decay steps at epochs $0, 20, 30$. Additionally, we used Automatic Mixed Precision, which is implemented by PyTorch. We initialize from a Kinetics-400 pre-trained model~\cite{kay2017kinetics}. 

\minisection{Regularization details} We use weight decay of $0.05$, a dropout~\cite{Hinton2012dropout} of $0.5$ before the final classification, dropout of $0.3$ after the \ocmodel~block, and DropConnect~\cite{Huang2016deepstochastic} with rate $0.2$.

\minisection{Training details} We use a standard crop size of $224$, and we jitter the scales from $256$ to $320$.

\minisection{Inference details} We take 3 spatial crops per single clip to form predictions over a single video in testing.

\section{Additional Model details}
\label{supp:model}

\subsection{Object-Region Attention}
\label{supp:model:ocmodel}

As explained in~\secref{sec:model:ocmodel}, there are two inputs to the {\ocmodel} block. The first is the output of the preceding transformer block, represented as a set of spatio-temporal tokens  $X\in \reals^{THW \times d}$. The second input is a set of bounding boxes for objects across time, denoted by $B\in \reals^{T O\times 4}$.


\minisection{Object-Region Attention} Given the patch token features $X$ and the boxes $B$, we use RoIAlign~\cite{maskrcnn2017iccv} layer, which uses the patch tokens $X$ and box coordinates $B$ to obtain object region crops. This is followed by max-pooling and an MLP. To these features we add a learnable object-time position encoding $\mathcal{P} \in \reals^{TO \times d}$ to encode the positional object information. We also use a coordinate embedding by applying an MLP on the boxes coordinates, resulting in $\mathcal{L} \in \reals^d$:
\begin{equation}
    \mathcal{L} := \text{MLP}(B)
\label{eq:supp:location_embd}
\end{equation}

where $B \in \reals^{T \times O \times d}$ is the boxes coordinates. This leads to an improved object features:
\begin{equation}
     \mathcal{O} := \text{MLP}(\text{MaxPool}(\text{RoIAlign}(X,B))) + \mathcal{L} + \mathcal{P}
\label{eq:supp:objects}
\end{equation}

where the token features are $X \in \reals^{THW \times d}$. We pass these features into the self-attention layers as explained in the ``Object-Region attention'' subsection in the main paper.





\input{Tables/lightweight_supp}

\section{Additional Ablations}
\label{supp:ablations}


We perform an ablation study of each of the components in~\tabref{supp:tab:ablations} to show the effectiveness of the different components of our model. All ablations are on the SomethingElse~\cite{materzynska2019something} dataset and we use Mformer ({\Mformer}) as the baseline architecture for {\ocmodel} unless stated otherwise. We also note that we refer to the ``Object-Dynamics Module'' as ODM stream.

\minisection{Contribution of appearance and motion streams} In~\tabref{supp:tab:abl_streams}, we show the ``Object-Region Attention'' is an important factor for the improvement, responsible for a $7.2\%$ gain improvement, with less than $2\%$ additional parameters over {\Mformer} (only 2M parameters addition compared to the baseline). This highlights our contribution that object interactions are indeed crucial for video transformers. Additionally, adding trajectory information with coordinates in the ``Object-Dynamics Module'' (ODM) improved by another $2.3\%$ but with a cost of $36\%$ additional parameters. We show later (see in~\Secref{supp:exper:lightweight}) that we can reduce the ODM size with smaller dimensions.

\minisection{\ocmodel~blocks} In~\tabref{supp:tab:abl_layer12_all}, we show which layers are most important for adding the \ocmodel~block. The experiments show that adding this information at the network's beginning, middle, and end is the most effective (layer 2, 7, 11). This experiment demonstrates that it is important to fuse object-centric representations starting from early layers and propagate them into the transformer-layers, thus affecting the spatio-temporal representations throughout the network. 




\minisection{Different self-attention in ``Object-Region Attention''} In~\tabref{supp:tab:abl_appearance}, we compared different self-attention mechanisms (as defined in~\cite{patrick2021keeping}): joint space-time, divided space-time, and trajectory attention to the \textit{\Mformer} baseline, which uses trajectory attention in all layers. We observed that trajectory attention is slightly better. However, it can be seen that our object region approach is not sensitive to these choices, indicating that the generic approach is the main reason for the observed improvements.

\minisection{Replacing {\ocmodel} with Trajectory Attention} We observe that joint and divided self-attention layers~\cite{gberta_2021_ICML,arnab2021vivit} have similar results to the trajectory attention~\cite{patrick2021keeping}, as seen in~\tabref{supp:tab:abl_appearance}. However, we would like to demonstrate that trajectory attention is not the main reason for the improvement when using {\ocmodel} with TimeSformer~\cite{gberta_2021_ICML} or MViT~\cite{mvit2021}. Thus, we replace our {\ocmodel} with a standard trajectory attention on the Diving48 and AVA datasets. The top1 accuracy on Diving48 are improved by $4.5\%$ (from $80.0$ to $84.5$) with trajectory attention, while using our \textit{\ocmodel+TimeSformer} achieves $88.0$ ($3.5\%$ improvements on top of that). The MAP on AVA are the same as the baseline with trajectory attention ($25.5$), while using our \textit{\ocmodel+MViT-B} achieves $26.6$ ($1.1$ improvements on top of the baseline). We note that our {\Mformer} is the baseline on EK100, SSv2, and SomethingElse, and therefore the trajectory attention is already part of the model, and hence this demonstration is not needed.


\minisection{Processing trajectory information} In~\tabref{supp:tab:abl_motion}, we compared our self-attention (see ``Object-Dynamics Module'' in~\Secref{sec:model:ocmodel}) with other standard baseline models: GCN~\cite{kipf2016semi} and trajectory self-attention~\cite{patrick2021keeping}. For the GCN, we use a standard implementation with 2 hidden layers, while for the trajectory attention, we treat the $O$ objects as the spatial dimension. We can see that self-attention is slightly better than trajectory self-attention (+$0.3\%$) and significantly better than GCN (+$2.0\%$). 




\minisection{Components on Diving48} Following our components ablations in~\tabref{tab:pitch}, we also validate our hypothesis on the Diving48 dataset. We used the TimeSformer\cite{gberta_2021_ICML} trained on videos of $32$ frames as a baseline for a fair comparison. It can be seen that a single layer version of the model already results in considerable improvement ($85.4\%$) and that it is important to apply it in the earlier layers of transformers than at the end ($86.8\%$ compared to $85.4\%$). Additionally, the ``Object-Dynamics Module'' improves performance to $87.5\%$. Finally, multiple applications of the layer further improve performance to $88.0\%$.

\minisection{Box Position Encoder} Our ``Box Position Encoder'' transforms from a tensor of size $TO$ to  size $THW$. Our implementation of this transformation uses box information so that each object is mapped to the ``correct'' region in space. A simpler approach would have been to expand the shape of $TO$ to $THW$ without using boxes. We refer to the latter as a standard tensor expansion. Comparing the two methods, we find out that our approach obtains $69.7$ compared to $68.4$, showing that our box-based encoding performs better.

\minisection{Combine the two streams in ORViT Block} As part of the ORViT block, we examined other operations to combine the two streams of information. We explored the following methods: element-wise multiplication, gating (with conv), and our simple sum. The results on SomethingElse are $68.8$, $68.7$, and $69.7$, respectively. In this case, our simple sum is superior to the other methods.

\minisection{$T \times O$ learnable embeddings} We observe a small difference when experimenting with $T \times O$ and separate $T$ and $O$ embeddings (69.6 Vs. 69.7) on the SomethingElse dataset.

\section{Additional Experiments}
\label{supp:exper}

Here we present additional experiments, including demonstrating a lightweight version of the ``Object-Dynamics Module'' that significantly reduces the model parameters without losing significant performance and complete results on the standard action recognition task.


\subsection{Light-weight {\ocmodel}}
\label{supp:exper:lightweight}


In~\tabref{supp:tab:abl_streams}, we show that the ``Object-Dynamics Module'' improves by $2.3\%$ the top-1 accuracy with an additional 39M parameters (148M Vs. 109M). We would like to demonstrate that model size can be significantly decreased, incurring a small performance loss. Most the parameters added by {\ocmodel} over the baseline {\Mformer} are in the ODM, and thus it is possible to use a smaller embedding dimension in ODM. Here, we present a light-weight version of the module that reduces the embedding dimensions without losing significant accuracy. See~\tabref{tab:lightweight}. 


As mentioned in the main paper (see~\Secref{sec:model}), we use $\widetilde{B}$ for the coordinate embeddings in the ``Object-Dynamics Module''. We observe that reducing the dimension of the coordinate embeddings ($\widetilde{B}$) from $768$ to $256$ has little impact on the action accuracy in SSv2 ($67.9\%$ Vs. $67.3\%$) and SomethingElse ($69.7\%$ Vs. $68.9\%$), although having only 114M model parameters (an addition of 5M parameters to the {\Mformer} baseline that has 109M). Indeed this indicates that our main approach is the main reason for the observed improvements and not necessarily the addition of parameters.


\subsection{Standard Action Recognition Results}
\label{supp:exper:results}

We next report in~\tabref{supp:table:ar} the full results table for the standard action recognition task, including extra models and details, which were not included in the main paper. 

Additionally, we add a light version of {\ocmodel} for each dataset. This version use embedding dimension of 256 in the ``Object-Dynamics Module'', as stated in~\Secref{supp:exper:lightweight}. In SSv2, the {\ocmodel}-Light model improves the {\Mformer} baseline by $0.8$ at the cost of additional $5M$ parameters ($5\%$ more parameters), while the {\ocmodel} model (non-light version) improves by $1.4\%$ at the cost of additional $39M$ parameters ($36\%$ more parameters). In Diving48, the {\ocmodel}-Light model improves the TimeSformer baseline by $6.8$ at the cost of additional $5M$ parameters ($3\%$ more parameters), while the {\ocmodel} model (non-light version) improves by $8\%$ at the cost of additional $39M$ parameters ($32\%$ more parameters). In EK100, the {\ocmodel}-Light model improves the {\Mformer}-HR baseline by $1.6, 1.5, 0.2$ (A, V, N) at the cost of additional $5M$ parameters ($5\%$ more parameters), while the {\ocmodel} model (non-light version) improves by $1.2, 1.4, 0.2$ at the cost of additional $39M$ parameters ($36\%$ more parameters). We note that the {\ocmodel}-Light even outperforms the non-light version (1.6 Vs. 1.2), demonstrating the object movement is less significant in this data.

Last, we separately evaluated the accuracy on Kinetics-400 with ORViT MViT-B 16x4 and noticed that it improved by +2.0\% over MViT-B 16x4.


\begin{table*}[tb!]
    \caption{\textbf{Comparison to the state-of-the-art on video action recognition.} We report pretrain, param ($10^6$), GFLOPS ($10^9$) and top-1 ($\%$) and top-5 ($\%$) video action recognition accuracy on SSv2. On Epic-Kitchens100 (EK100), we report top-1 ($\%$) action (A), verb (V), and noun (N) accuracy. On Diving48 we report pretrain, number of frames, param ($10^6$) and top-1 ($\%$) video action recognition accuracy. Difference between baselines (MF/MF-L for SSv2, TimeSformer for Diving48, MF-HR for EK100) and {\ocmodel} is denoted by ({$+$X}). We denote methods that do not use bounding boxes with \protect\methodwithoutboxes.}
    
    \begin{subtable}[t]{.99\linewidth}
	\tablestyle{15.5pt}{1.05}
	\scriptsize
	    \centering
	    \setlength{\tabcolsep}{3pt}
	    \caption{\bf{Something--Something V2}}
		\input{Tables/sota_ssv2_supp}

		\label{supp:tab:sota_ssv2}
	\end{subtable}
	\\
    
    \begin{subtable}[t]{.48\linewidth}
    \tablestyle{1.5pt}{1.05}
    \scriptsize
        \centering
        \caption{\bf{Diving48}}
        \label{supp:tab:sota_diving48}
		\input{Tables/sota_diving48_supp}
	\end{subtable}
	\hfill
	\begin{subtable}[t]{.53\linewidth}
	\tablestyle{1.5pt}{1.05}
	\scriptsize
	    \centering
	    \caption{\bf{Epic-Kitchens100}}
	    \label{supp:tab:sota_ek100}
	    \input{Tables/sota_epic_kitchens_supp}
	\end{subtable}
    \label{supp:table:ar}
\end{table*}

\section{Qualitative Visualizations}
\label{supp:qual}
\begin{figure*}[t!]
    \centering
    \includegraphics[width=\linewidth]{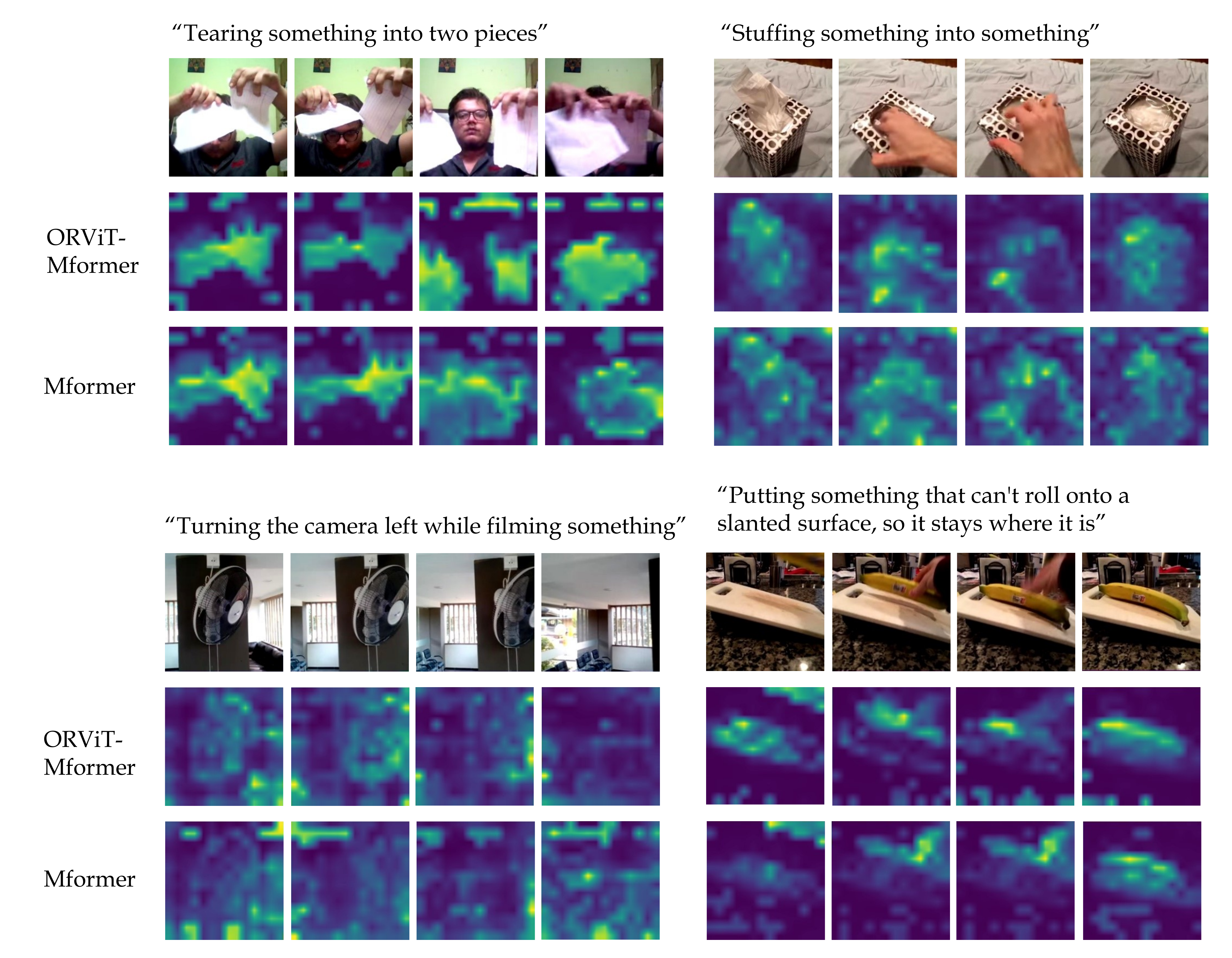}
    \caption{\textbf{Attention Maps} comparison between the \textit{\ocmodel+\Mformer} and the \textit{\Mformer} on videos from the SSv2 dataset. The visualization shows the attention maps corresponding to the CLS query.}
    \label{fig:vis1}
\end{figure*}

\begin{figure*}[t!]
    \centering
    \includegraphics[width=\linewidth]{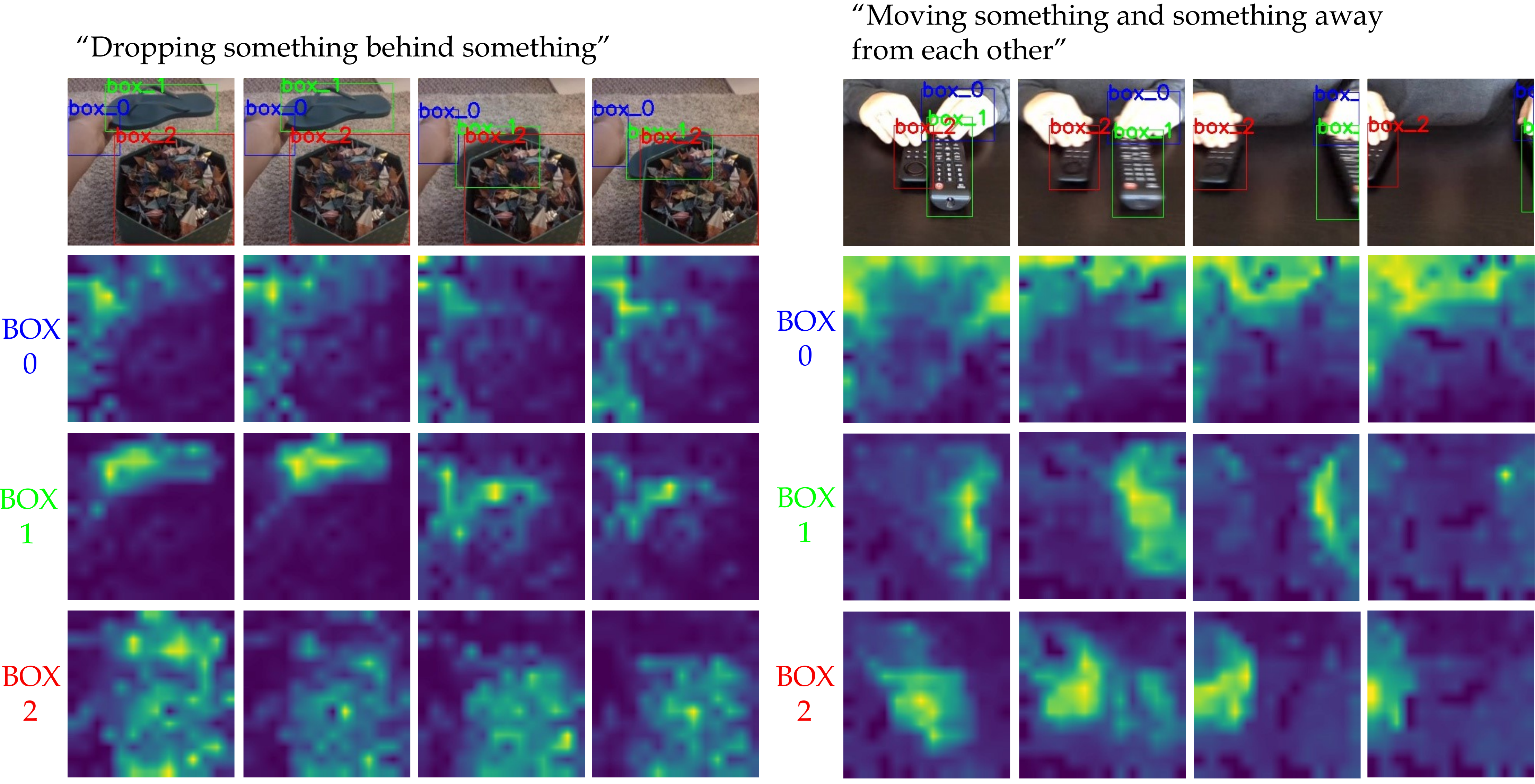}
    \caption{\textbf{Object contribution to the patch tokens.} For each object token, we plot the attention weight given by the patch tokens, normalized over the patch tokens.}
    \label{fig:vis2}
\end{figure*}

To provide insight into the inner representation of {\ocmodel} we provide further visualization next. 
See~\figgref{fig:vis1} and~\figgref{fig:vis2}. In~\figgref{fig:vis1}, we visualize the attention map of the CLS token on all spatial tokens. It can be seen that object-regions indeed affect these spatial maps. For example, ``Tearing something into two pieces'' (top left corner) demonstrates that \textit{\ocmodel+\Mformer} successfully separates the two pieces of the paper, while the \textit{\Mformer} baseline does not. Next, in~\figgref{fig:vis2} we visualize the attention allocated to each of the object keys. It can be seen that the object keys in \textit{\ocmodel} indeed affect their corresponding spatial tokens.

\renewcommand{\thefootnote}{\arabic{footnote}}

%% file: Tables/ablations_supp.tex
\begin{table*}[tb!]
	\begin{subtable}[t]{.53\linewidth}
	\tablestyle{5.5pt}{1.05}
	    \centering
	    \caption{\bf{Streams}}
	    \label{supp:tab:abl_streams}
	    \input{Tables/abl_streams}
	\end{subtable}
    \quad
    \begin{subtable}[t]{.41\linewidth}
    \tablestyle{5.5pt}{1.05}
        \centering
        \caption{\bf{Blocks Ablation}}
        \label{supp:tab:abl_layer12_all}
		\input{Tables/abl_layer12_all}
	\end{subtable}
	\\
	\vspace{-0.8em}

    \begin{subtable}[t]{.32\linewidth}
    \tablestyle{5.5pt}{1.05}
        \centering
        \caption{\bf{Object-Region Attention}}
        \label{supp:tab:abl_appearance}
		\input{Tables/abl_appearance}
	\end{subtable}
    \quad
    \begin{subtable}[t]{.27\linewidth}
    \tablestyle{6.5pt}{1.05}
        \centering
        \caption{\bf{Object-Dynamic Module}}
        \label{supp:tab:abl_motion}
		\input{Tables/abl_motion}
	\end{subtable}
    \quad
    \begin{subtable}[t]{.35\linewidth}
    \tablestyle{5.5pt}{1.0}
        \centering
        \caption{\bf{Components}}
        \label{supp:tab:pitch_diving}
		\input{Tables/abl_pitch_supp}
	\end{subtable}

    \caption{\textbf{Ablations.} Evaluation of different model ablations and baselines on the ``SomethingElse'' split (Tables (a-d) see text). We report pretrain, param ($10^6$), GFLOPS ($10^9$), and top-1 and top-5 video action recognition accuracy. Table (e) reports ablations on the Diving dataset.}
	\vspace{-0.8em}
    \label{supp:tab:ablations}
\end{table*}

%% file: Tables/abl_streams.tex
\begin{tabularx}
{\linewidth}{@{}l c c l@{} l@{} c c}
    \toprule
    {Model} & {Top-1} & {Top-5} & {GFLOPs ($10^9$)~~~~} & {Param ($10^6$)}\\ 
    \midrule
    Baseline & 60.2 & 85.5 & \reldif~1 (369.5) & \reldif~1 (109) \\
    + Object-Region Attention & 67.4 & 89.8 & \reldif~1.03 (382) & \reldif~1.02 (111) \\
    + Object-Dynamics Module & \bf 69.7 & \bf 91.0 & \reldif~1.1 (405) & \reldif~1.36 (148) \\
    \bottomrule
\end{tabularx}

%% file: Tables/abl_layer12_all.tex
\begin{tabularx}
{\linewidth}{@{}l c c l@{} l@{}}
    \toprule
    {Layers} & {Top-1} & {Top-5} & {GFLOPs ($10^9$)~~~~} & Param ($10^6$)\\ 
    \midrule
    Baseline & 60.2 & 85.8 & \reldif~1 (369.5) & \reldif~1 (109) \\ 
	2 & 68.9 & 90.4 & \reldif~1.03 (381) & \reldif~1.12 (122) \\ 
	7 & 67.8 & 89.5 & \reldif~1.03 (381) & \reldif~1.12 (122) \\ 
	11 & 66.8 & 89.3 & \reldif~1.03 (381) & \reldif~1.12 (122) \\
	2, 7 & 69.3 & 90.6 & \reldif~1.06 (393) & \reldif~1.24 (135) \\ 
	2, 7, 11 & \textbf{69.7} & \textbf{91.0} & \reldif~1.1 (405) & \reldif~1.36 (148) \\ 
    \bottomrule
\end{tabularx}

%% file: Tables/abl_appearance.tex
\begin{tabularx}
{\linewidth}{@{}l c c r@{} r}
    \toprule
    {Model} & {Top-1} & {Top-5} \\ 
    \midrule
    Baseline & 60.2 & 85.8 \\
    Ours /\w~Joint attention & 68.9 & 90.4 \\
    Ours /\w~Divided attention & 69.3 & 90.6 \\
    Ours /\w~Trajectory attention & \bf 69.7 & \bf 91.0 \\
    \bottomrule
\end{tabularx}

%% file: Tables/abl_motion.tex
\begin{tabularx}
{\linewidth}{@{}l c c r@{} r}
    \toprule
    {Model} & {Top-1} & {Top-5} \\ 
    \midrule
    GCN & 67.7 & 89.8 \\
    Trajectory attention & 69.4 & 90.5 \\
    Self-attention & \bf 69.7 & \bf 91.0 \\
    \bottomrule
\end{tabularx}

%% file: Tables/abl_pitch_supp.tex
\begin{tabularx}
{\linewidth}{@{}l c l@{} c c}
    \toprule
    {Layers} & {Top-1} & Param \\ 
    \midrule
    Baseline & 80.0 & \reldif~1 (121) \\ 
	{\ocmodel} [:12] & 85.4 & \reldif~1.01 (122) \\ 
	{\ocmodel} [:2] & 86.8 & \reldif~1.01 (122) \\ 
	{\ocmodel} [:2] + ODM & 87.5 & \reldif~1.11 (134) \\ 
	{\ocmodel} [:2, 7, 11] + ODM & \textbf{88.0} & \reldif~1.32 (160) \\ 
    \bottomrule
\end{tabularx}

%% file: Tables/lightweight_supp.tex
\begin{table*}[tb!]
	\centering
	\tablestyle{10.5pt}{1.15}
	\begin{tabular}{l | c | c | c | c | l@{} | l@{} }
        \toprule
        {Dataset} & {Model}  & {ODM Dimension} & {Top-1} & {Top-5} & {GFLOPs} $(10^9)$~~~ & {Param} $(10^6)$~~~ \\ 
        \midrule
        \multirow{4}{*}{SomethingElse} & {Baseline} & - & 60.2 & 85.8 & \reldif~1 (369.5) & \reldif~1 (109)\\
        \cline{2-7}
         & \multirow{3}{*}{{\ocmodel}} & 128 & 68.7 & 90.3 & \reldif~1.03 (382) & \reldif~1.03 (112)\\
         & & 256 & 68.9 & 90.5 & \reldif~1.04 (383) & \reldif~1.05 (114)\\
         & & 768 & 69.7 & 91.0 & \reldif~1.1 (405) & \reldif~1.36 (148)\\
        \midrule
        \multirow{4}{*}{SSv2} & {Baseline} & - & 66.5 & 90.1 & \reldif~1 (369.5) & \reldif~1 (109)\\
        \cline{2-7}
        & \multirow{3}{*}{{\ocmodel}}  & 128 & 67.2 & 90.4 & \reldif~1.03 (382) & \reldif~1.03 (112)\\
        & & 256 & 67.3 & 90.5 & \reldif~1.04 (383) & \reldif~1.05 (114)\\
        & & 768 & 67.9 & 90.5 & \reldif~1.1 (405) & \reldif~1.36 (148)\\
	\end{tabular}
	\caption{\textbf{A light-weight version of~\ocmodel}.}
	\label{tab:lightweight}
	\vspace{-1.0em}
\end{table*}


%% file: Tables/sota_ssv2_supp.tex
\renewcommand{\thefootnote}{\fnsymbol{footnote}}
\begin{tabularx}
{0.68\linewidth}{@{}l c c c r@{} r@{}}
    \toprule
    {Model} & Pretrain & {Top-1} & {Top-5} & {GFLOPs$\times$views} ($10^9$) & {~~~~Param} ($10^6$)\\ 
    \midrule
    SlowFast, R50\methodwithoutboxes & K400 & 61.7 & 87.0 & 65.7$\times$3$\times$1 & 34.1 \\
    SlowFast, R101\methodwithoutboxes & K400 & 63.1 & 87.6 & 106$\times$3$\times$1 & 53.3 \\
    TSM\methodwithoutboxes & K400 & 63.4 & 88.5 & 62.4$\times$3$\times$2 & 42.9\\
    STM\methodwithoutboxes & IN-1K & 64.2 & 89.8 & 66.5$\times$3$\times$10 & - \\
    MSNet\methodwithoutboxes & IN-1K & 64.7 & 89.4 & 67$\times$1$\times$1 & 24.6 \\
    TEA\methodwithoutboxes & IN-1K & 65.1 & - & 70$\times$3$\times$10 & - \\
    bLVNet\methodwithoutboxes & IN-1K & 65.2 & 90.3 & 128.6$\times$3$\times$10 & - \\
    \midrule
    VidTr-L\methodwithoutboxes & { IN-21K+K400} &  60.2 & - & 351$\times$3$\times$10 & - \\ 
    TimeSformer-L\methodwithoutboxes & IN-21K & 62.5 & - & 1703$\times$3$\times$1 & 121.4 \\
    ViViT-L\methodwithoutboxes & { IN-21K+K400}  & 65.4 & 89.8 & 3992$\times$4$\times$3 & -\\
    MViT-B, 32\methodwithoutboxes & K400  &  67.1 &  90.8 & 170$\times$3$\times$1 & 36.6 \\
    MViT-B, 64\methodwithoutboxes & K400  &  67.7 &  90.9 & 455$\times$3$\times$1 & 36.6 \\
    MViT-B, 32\methodwithoutboxes & K600  &  67.8 &  91.3 & 170$\times$3$\times$1 & 36.6 \\
    MViT-B, 64\methodwithoutboxes & K600  &  68.7 &  91.5 & 236$\times$3$\times$1 & 53.2 \\

    \midrule
    \Mformer\methodwithoutboxes & { IN-21K+K400} & 66.5 & 90.1 &$369.5\times3\times1$ & 109\\
    \Mformer-L\methodwithoutboxes & { IN-21K+K400} & 68.1 & 91.2 &$1185.1\times3 \times1$ & 109 \\
    \midrule
    \Mformer+STRG\methodwithboxes & {\scriptsize IN+K400}  & 66.1  & 90.0 & $375\times3 \times1$  & 117 \\
    \Mformer+STIN\methodwithboxes & {\scriptsize IN+K400}  & 66.5 & 89.8 & $375.5\times3 \times1$  & 111 \\
    \Mformer+STRG+STIN\methodwithboxes & {\scriptsize IN+K400}  & 66.6  & 90.0 & $375.5\times3 \times1$  & 119 \\
    \midrule
    \bf \ocmodel~\Mformer-Light\methodwithboxes~(Ours) & { IN-21K+K400}  & \textbf{67.3}\gcol{$+$0.8} & 90.5\gcol{$+$0.4} & $383\times3 \times1$ & 114\bcol{$+$5\%}\\
    \bf \ocmodel~\Mformer\methodwithboxes~(Ours) & { IN-21K+K400}  & \textbf{67.9}\gcol{$+$1.4} & 90.5\gcol{$+$0.4} & $405\times3 \times1$ & 148 \bcol{$+$36\%}\\
    \bf \ocmodel~\Mformer-L\methodwithboxes~(Ours) & { IN-21K+K400}  & \bf 69.5\gcol{$+$1.4} & \bf 91.5\gcol{$+$0.3} & $ 1259\times3 \times1$ & 148 \bcol{$+$36\%}\\

    \bottomrule
 \end{tabularx}
\renewcommand{\thefootnote}{\arabic{footnote}}

%% file: Tables/sota_diving48_supp.tex
\renewcommand{\thefootnote}{\fnsymbol{footnote}}
\begin{tabularx}
{\linewidth}{@{}l c c c c r}
    \toprule
    {Model} & {Pretrain} & {Frames} & {Top-1} & {Params ($10^6$)} \\ 
    \midrule
    I3D\methodwithoutboxes & K400 & 8 & 48.3 & - \\
    TSM\methodwithoutboxes & ImageNet & 3 & 51.1 & 42.9 \\
    TSN\methodwithoutboxes & ImageNet & 3 & 52.5 & -\\
    GST-50\methodwithoutboxes & ImageNet & 8 & 78.9 & - \\
    ST-S3D\methodwithoutboxes & K400 & 8 & 50.6 & -\\

    \midrule
    SlowFast, R101\methodwithoutboxes & K400 & 16 & 77.6 & 53.3 \\
    TimeSformer\methodwithoutboxes & IN-21K & 16 & 74.9 & 121 \\
    TimeSformer-HR\methodwithoutboxes & IN-21K & 16 & 78.0 & 121 \\
    TimeSformer-L\methodwithoutboxes & IN-21K & 96 & 81.0 & 121\\
    TQN\methodwithoutboxes & K400 & ALL & 81.8 & - \\

    \midrule
    TimeSformer\methodwithoutboxes & IN-21K & 32 & 80.0 & 121 \\
    TimeSformer + STIN\methodwithboxes & IN-21K & 32 & 81.0 & 123\\
    TimeSformer + STRG\methodwithboxes & IN-21K & 32 & 78.1 & 129 \\
    TimeSformer + STRG + STIN\methodwithboxes & IN-21K & 32 & 83.5 & 132 \\
    
    \midrule
    \bf \ocmodel~TimeSformer-Light\methodwithboxes~(Ours) & {\scriptsize IN-21K} & 32 & \bf 86.8\gcol{$+$6.8} & 126\bcol{$+$3\%} \\
    \bf \ocmodel~TimeSformer\methodwithboxes~(Ours) & {\scriptsize IN-21K} & 32 & \bf 88.0\gcol{$+$8.0} & 160\bcol{$+$32$\%$} \\
    \bottomrule
\end{tabularx}
\renewcommand{\thefootnote}{\arabic{footnote}}



    

%% file: Tables/sota_epic_kitchens_supp.tex
\renewcommand{\thefootnote}{\fnsymbol{footnote}}
\begin{tabularx}
    {\linewidth}{@{}l c c c c c}
    \toprule
    {Method} & Pretrain & {A} & {V} & {N} & {Params ($10^6)$} \\ 
    \midrule
    TSN\methodwithoutboxes & IN-1K & 33.2 & 60.2 & 46.0 & - \\ 
    TRN\methodwithoutboxes & IN-1K & 35.3 & 65.9 & 45.4 & - \\ 
    TBN\methodwithoutboxes & IN-1K & 36.7 & 66.0 & 47.2 & - \\
    TSM\methodwithoutboxes & IN-1K & 38.3 & 67.9 & 49.0 & - \\
    SlowFast\methodwithoutboxes & K400  & 38.5 & 65.6 & 50.0 & - \\
    
    \midrule
    TimeSformer\methodwithoutboxes & IN-21K & 32.9 & 55.8 & 50.1 & 121 \\
    ViViT-L & { IN-21K+K400}  & 44.0 & 66.4 & 56.8 & - \\
    \Mformer\methodwithoutboxes & { IN-21K+K400} & 43.1 & 66.7 & 56.5 & 109 \\
    \Mformer-L\methodwithoutboxes & { IN-21K+K400} & 44.1 & 67.1 & 57.6 & 109 \\
    \Mformer-HR\methodwithoutboxes & { IN-21K+K400} & 44.5 & 67.0 & 58.5 & 109 \\

    \midrule
    \Mformer-HR + STIN & { IN-21K+K400}  & 44.2 & 67.0 & 57.9 & 111 \\
    \Mformer-HR + STRG & { IN-21K+K400}  & 42.5 & 65.8 & 55.4 & 117 \\
    MF-HR + STRG + STIN & { IN-21K+K400}  & 44.1 & 66.9 & 57.8 & 119 \\    
    \midrule
    \bf \ocmodel~\Mformer-HR & { IN21K+K400} & \bf  45.7\gcol{$+$1.2} & \bf  68.4\gcol{$+$1.4} & \bf  58.7\gcol{$+$.2} & 148\bcol{$+$36\%} \\
    \bf \ocmodel~\Mformer-HR-Light & { IN21K+K400} & \bf  46.1\gcol{$+$1.6} & \bf  68.5\gcol{$+$1.5} & \bf  58.7\gcol{$+$.2} & 114\bcol{$+$5\%} \\
    \bottomrule
\end{tabularx}
\renewcommand{\thefootnote}{\arabic{footnote}}
